\documentclass{article}

\usepackage{microtype}
\usepackage{graphicx}
\usepackage{booktabs} 
\usepackage[noend]{algorithmic}

\usepackage{hyperref}

\usepackage[accepted]{icml2020}

\icmltitlerunning{\acro{BINOCULARS} for Efficient, Nonmyopic Sequential Experimental Design}

\usepackage[utf8]{inputenc} 
\usepackage[T1]{fontenc}    
\usepackage{url}            
\usepackage{booktabs}       
\usepackage{amsfonts}       
\usepackage{nicefrac}       

\usepackage{breakurl}
\usepackage[font=small]{caption}
\usepackage{wrapfig}
\usepackage{textcomp}
\usepackage{blindtext}

\usepackage{multirow}
\usepackage{dsfont}
\usepackage{multicol}
\usepackage{amsmath}
\usepackage{amsthm}
\usepackage{amssymb}
\usepackage{bm}
\usepackage{mathrsfs}
\usepackage{xcolor}
\usepackage{xspace}
\usepackage{colortbl}
\usepackage[labelformat=simple]{subcaption}
\usepackage[hang,flushmargin]{footmisc}

\usepackage{color}
\usepackage{mathtools}
\usepackage{soul}

\usepackage{pgfplots}
\pgfplotsset{
  compat=newest,
  plot coordinates/math parser=false,
  tick label style={font=\footnotesize, /pgf/number format/fixed},
  label style={font=\small},
  legend style={font=\small},
  every axis/.append style={
    tick align=outside,
    clip mode=individual,
    scaled ticks=false,
    thick,
    tick style={semithick, black}
  }
}

\pgfkeys{/pgf/number format/.cd, set thousands separator={\,}}

\usetikzlibrary{calc}
\usepgfplotslibrary{external}
\usetikzlibrary{external}
\newlength\figurewidth
\newlength\figureheight

\newlength\sbsfigurewidth
\newlength\sbsfigureheight

\newcommand{\mc}[1]{\mathcal{#1}}
\newcommand{\data}{\mc{D}}

\newcommand{\given}{\mid}
\newcommand{\E}{\mathbb{E}}
\newcommand{\intd}[1]{\,\mathrm{d}{#1}}

\newcommand{\acro}[1]{\textsc{\lowercase{#1}}}
\newcommand{\acrob}[1]{\textbf{\textsc{\lowercase{#1}}}}

\newcommand{\bo}{\acro{BO}\xspace}
\newcommand{\bq}{\acro{BQ}\xspace}

\newcommand{\sed}{\acro{SED}\xspace}
\newcommand{\gp}{\acro{GP}\xspace}
\newcommand{\dyp}{\acro{DP}\xspace}
\newcommand{\ei}{\acro{EI}\xspace}
\newcommand{\EI}{\acro{EI}\xspace}
\newcommand{\DPP}{\acro{DPP}\xspace}
\newcommand{\R}{\acro{R}\xspace}
\newcommand{\qei}{$q$-\acro{EI}\xspace}
\newcommand{\gap}{\acro{GAP}\xspace}

\newcommand{\lda}{\acro{LDA}\xspace}
\newcommand{\svm}{\acro{SVM}\xspace}
\newcommand{\unct}{\acro{UNCT}\xspace}
\newcommand{\binoc}{\acro{BINOCULARS}\xspace}
\newcommand{\glasses}{\acro{GLASSES}\xspace}
\DeclareMathOperator*{\argmax}{arg\,max}

\begin{document}

\twocolumn[
\icmltitle{\binoc for Efficient, Nonmyopic Sequential Experimental Design}



\icmlsetsymbol{equal}{*}

\begin{icmlauthorlist}
\icmlauthor{Shali Jiang}{equal,washu}
\icmlauthor{Henry Chai}{equal,washu}
\icmlauthor{Javier Gonzalez}{amz}
\icmlauthor{Roman Garnett}{washu}
\end{icmlauthorlist}

\icmlaffiliation{washu}{Department of Computer Science and Engineering, Washington University in Saint Louis, Saint Louis, Missouri, USA}
\icmlaffiliation{amz}{Amazon Research Cambridge, Cambridge, UK}

\icmlcorrespondingauthor{Shali Jiang}{jiang.s@wustl.edu}
\icmlcorrespondingauthor{Henry Chai}{hchai@wustl.edu}

\icmlkeywords{sequential experimental design, active learning, Bayesian optimization, Bayesian quadrature, Gaussian processes}

\vskip 0.3in
]



\printAffiliationsAndNotice{\icmlEqualContribution} 

\begin{abstract}

Finite-horizon sequential experimental design (\acro{SED}) arises naturally in many contexts, including hyperparameter tuning in machine learning among more traditional settings. Computing the optimal policy for such problems requires solving Bellman equations, which are generally intractable. Most existing work resorts to severely myopic approximations by limiting the decision horizon to only a single time-step, which can underweight exploration in favor of exploitation. We present \binoc: \acrob{B}atch-\acrob{I}nformed \acrob{NO}nmyopic \acrob{C}hoices, \acrob{U}sing \acrob{L}ong-horizons for \acrob{A}daptive, \acrob{R}apid \acrob{S}\acro{ED}, a general framework for deriving efficient, \emph{nonmyopic} approximations to the optimal experimental policy.  Our key idea is simple and surprisingly effective: we first compute a one-step optimal batch of experiments, then select a single point from this batch to evaluate. We realize \binoc for Bayesian optimization and Bayesian quadrature -- two notable \acro{SED} problems with radically different objectives -- and demonstrate that \binoc significantly outperforms myopic alternatives in real-world scenarios.
\end{abstract}

\section{Introduction}
Many real-world problems can be framed as finite-horizon sequential experimental design (\sed), wherein an agent adaptively designs a prespecified number of experiments seeking to maximize some data-dependent utility function. The optimal policy for \sed can be formulated as dynamic programming (\dyp), which balances the inherent tradeoff between exploitation (immediately advancing the goal) and exploration (learning for the future). However, this optimal policy is intractable even for simple problems \citep{powell_2011}. Common approximation schemes include rollout, Monte Carlo tree search \citep{bertsekas2017dynamic, powell_2011}, or simply artificially limiting the horizon, known as a \emph{myopic} approximation.

In this work, we propose a novel method for \emph{efficient and nonmyopic} \sed, called \binoc: \acrob{B}atch-\acrob{I}nformed \acrob{NO}nmyopic \acrob{C}hoices, \acrob{U}sing \acrob{L}ong-horizons for \acrob{A}daptive, \acrob{R}apid \acrob{S}\acro{ED}. 
\binoc is inspired by the fact that the optimal batch (or non-adaptive) design is an approximation to the optimal sequential (or adaptive) design. 
In fact, the optimal adaptive and non-adaptive designs are exactly the same in some notable cases where the data utility does not depend on the observed outcomes, such as maximizing information gain for a fixed Gaussian process (\gp) \citep{krause2007nonmyopic}.
Even when this is not the case, we show that the optimal batch expected utility is a lower bound of the optimal sequential expected utility. Furthermore, it is tighter than the one-step optimal policy's implied expected utility. 
Motivated by this insight, \binoc iteratively computes an optimal batch of designs, then chooses one point from this batch. 
While many existing methods construct batch policies by simulating a sequential policy \citep{ginsbourger2010kriging, desautels2014parallelizing, jiang2018efficient}, 
\binoc goes the other way and ``reduces'' sequential design to batch design.

\binoc is a general framework applicable to any \sed problem. 
In this paper, we realize this framework on two important yet \emph{fundamentally different} \sed tasks: Bayesian optimization (\bo) \citep{kushner1964new, movckus1975bayesian, shahriari2016taking} and Bayesian quadrature (\bq) \citep{larkin72, diaconis88, hagan91}. In \bo, an agent repeatedly queries an expensive function seeking its global optimum, whereas in \bq the goal is to estimate an intractable integral of the function. 

For both problems, many popular policies are myopic: examples include expected improvement (\ei) for \bo \citep{movckus1975bayesian} and uncertainty sampling (\unct) for \bq \citep{gunter14}. 
These are all one-step optimal for maximizing particular utility functions in expectation. While they are computationally efficient and give reasonably good empirical results, they are liable to suffer from myopia and over-exploitation. Nonmyopic alternatives have recently been applied to \bo \citep{gonzalez2016glasses, lam2016bayesian, yue2019why}, and although results are promising, these are typically costly to compute.

\begin{figure*}
	\centering
	\begin{subfigure}[b]{.65\columnwidth}
		\centering
		\includegraphics[width=0.65\columnwidth]{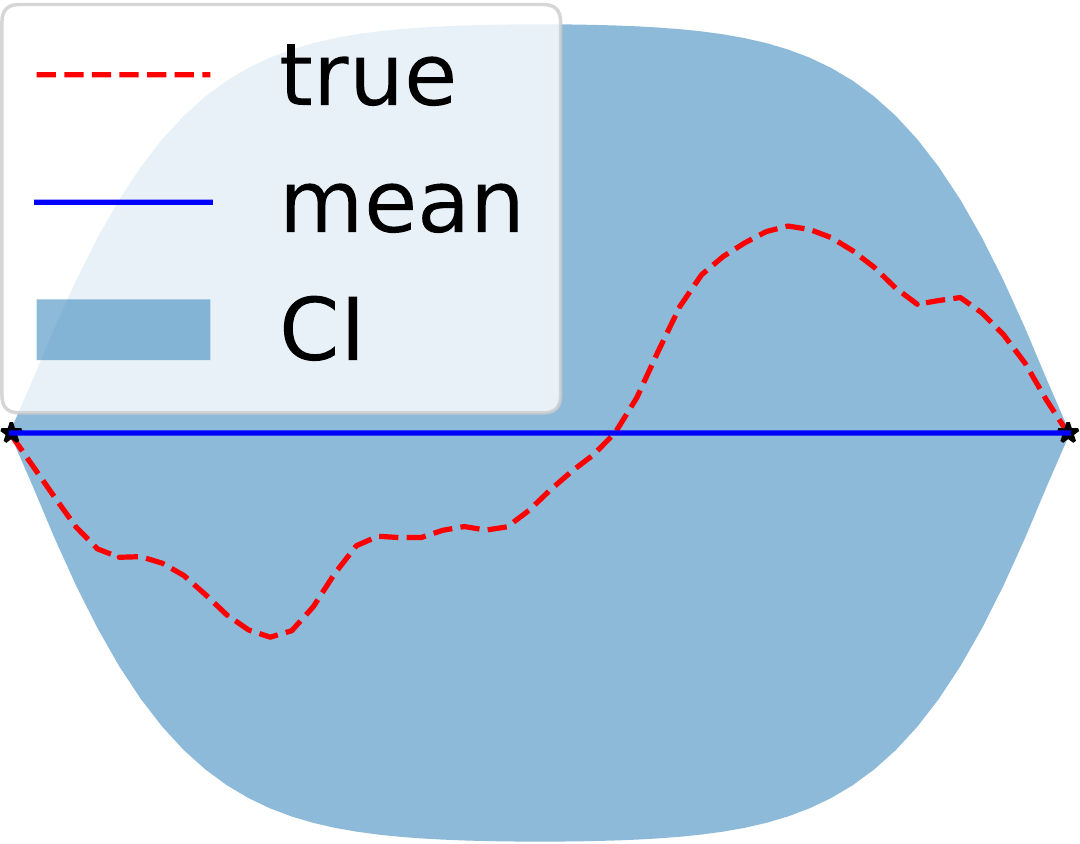}
		\subcaption{initial state}
		\label{fig:init}
	\end{subfigure}
	\begin{subfigure}[b]{.65\columnwidth}
		\centering
		\includegraphics[width=.65\columnwidth]{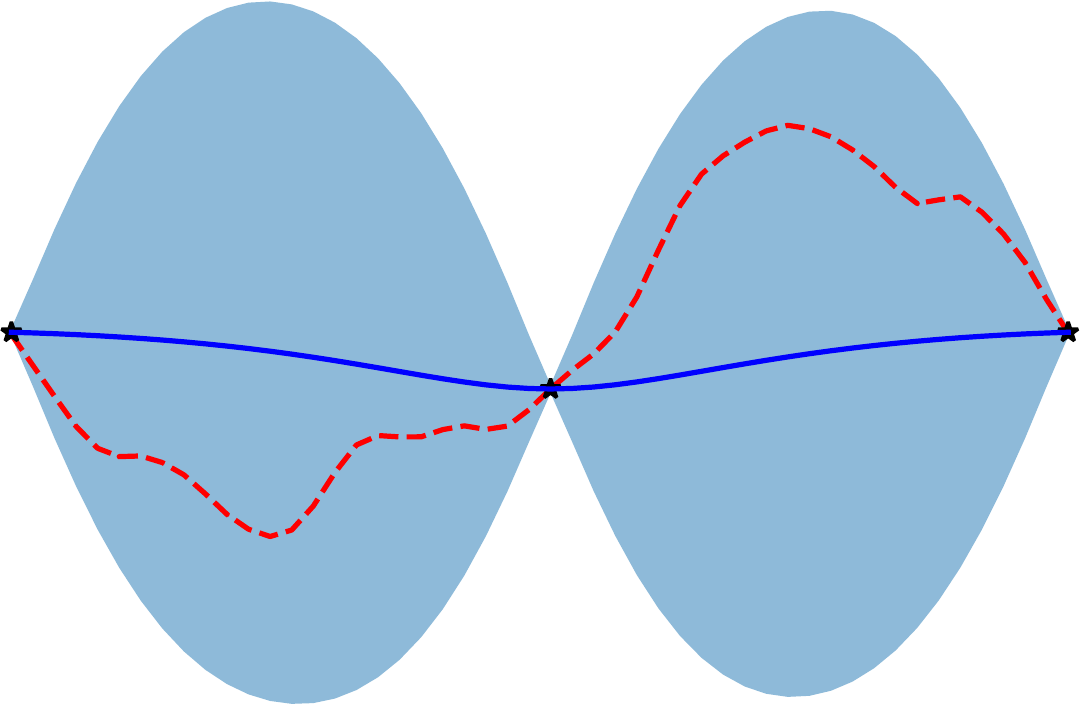}
		\subcaption{\ei iteration 1}
		\label{fig:ei1}
	\end{subfigure}
	\begin{subfigure}[b]{.65\columnwidth}
		\centering
		\includegraphics[width=.65\columnwidth]{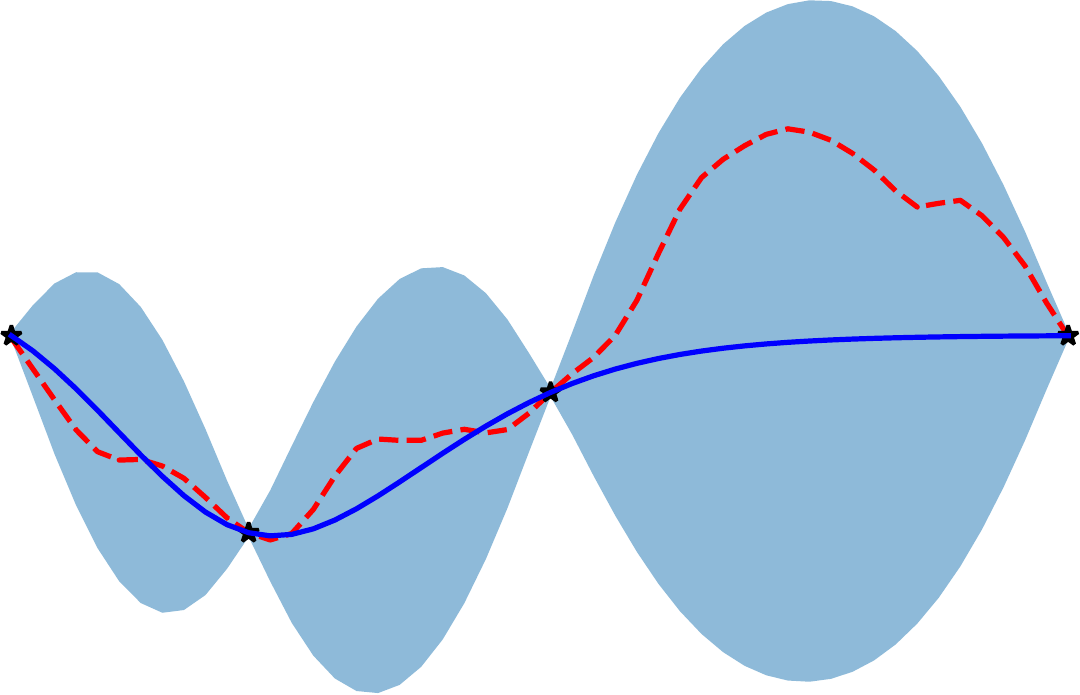}
		\subcaption{\ei iteration 2}
		\label{fig:ei2}
	\end{subfigure}
	\vspace{.5em}
	\begin{subfigure}[b]{.65\columnwidth}
		\centering
		\includegraphics[width=0.65\columnwidth]{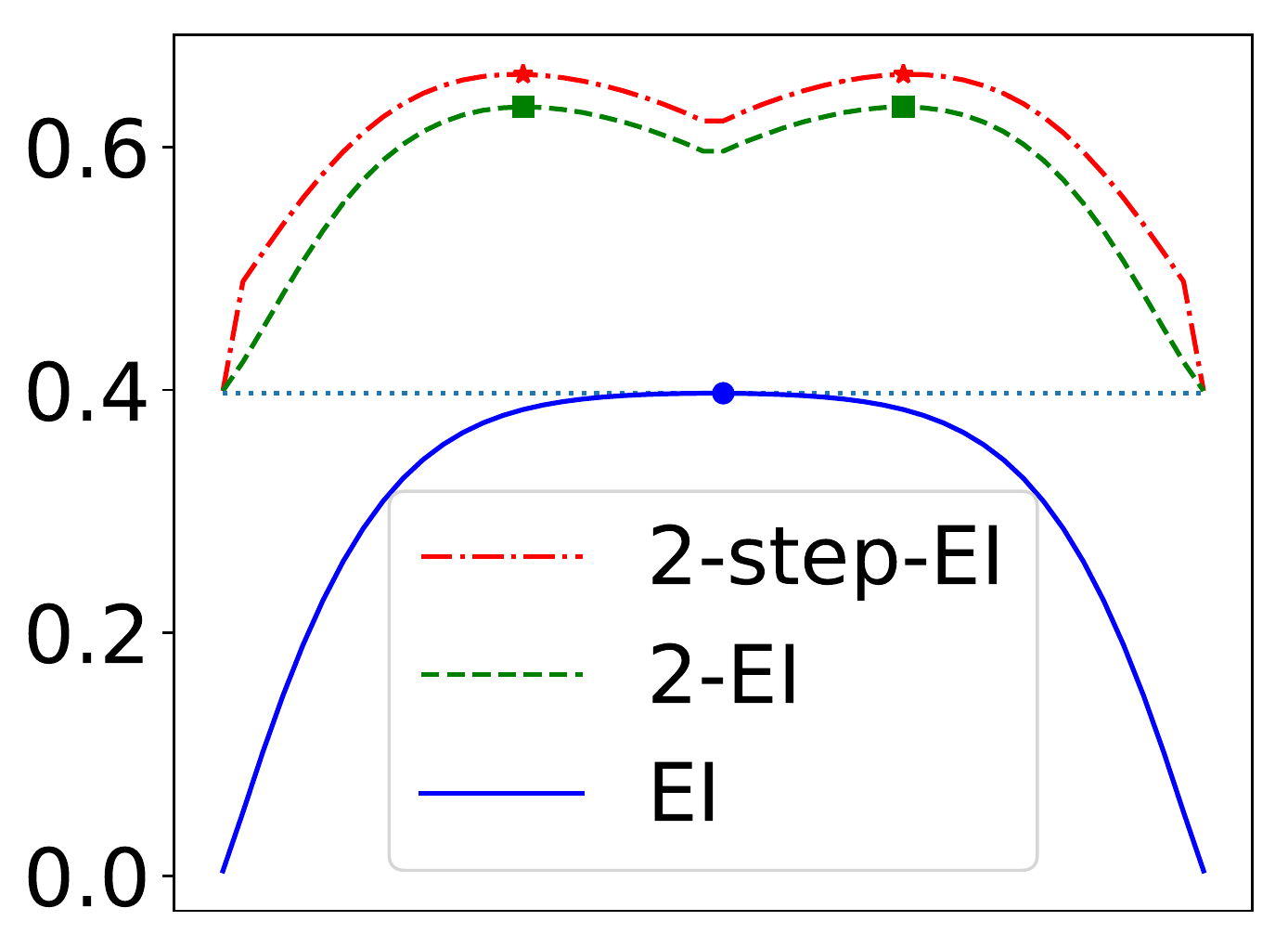}
		\subcaption{\ei, 2-\ei and 2-step-\ei}
		\label{fig:ei_vs_2ei}
	\end{subfigure}
	\begin{subfigure}[b]{.65\columnwidth}
		\centering
		\includegraphics[width=.65\columnwidth]{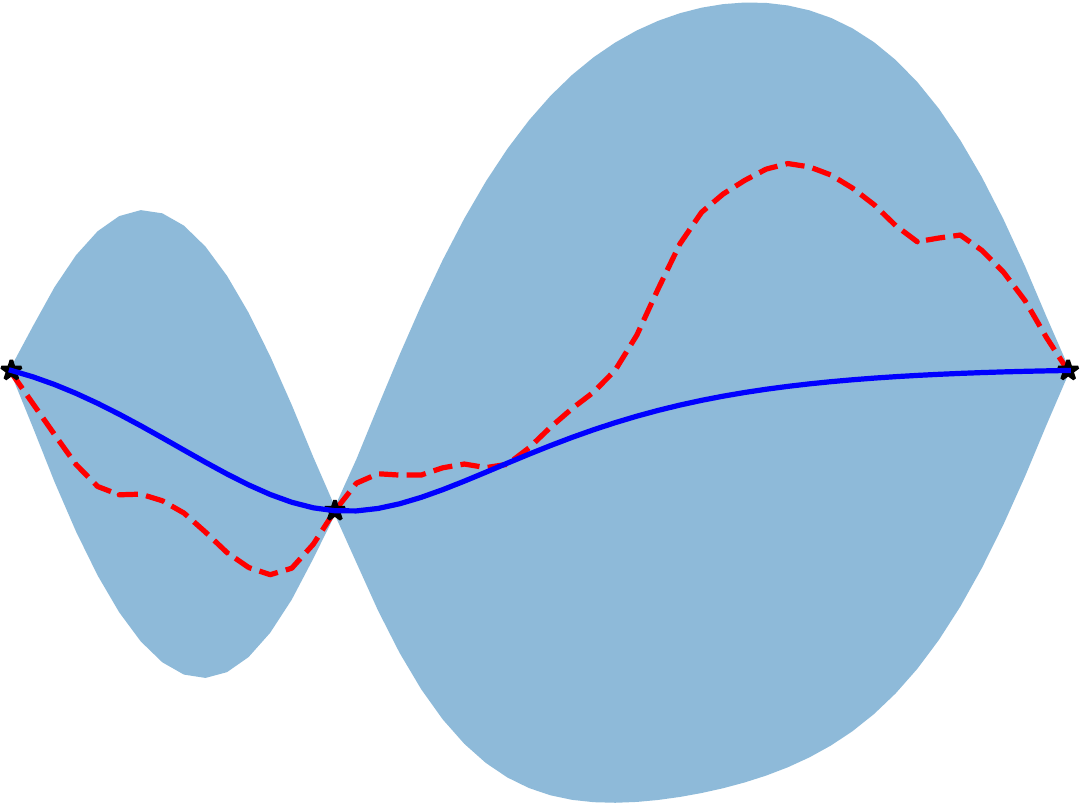}
		\subcaption{2-\ei iteration 1}
		\label{fig:2ei1}
	\end{subfigure}
	\begin{subfigure}[b]{.65\columnwidth}
		\centering
		\includegraphics[width=.65\columnwidth]{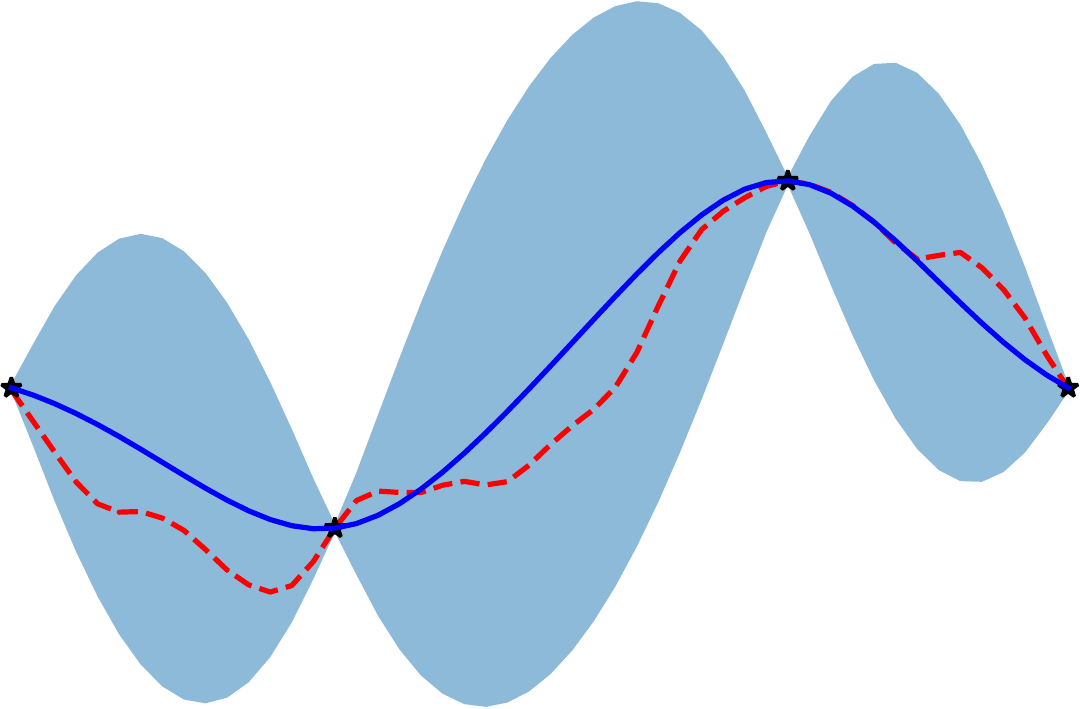}
		\subcaption{2-\ei iteration 2}
		\label{fig:2ei2}
	\end{subfigure}
	\caption{An illustration of our proposed nonmyopic method applied to \bo.
		(a) A function in $[-1,1]$ drawn from a \gp where the two end points are known to be zero. (b) and (c) show two iterations of \bo with the \ei acquisition function.
		(d) \ei, 2-\ei and $2$-step-\ei curves with their respective maximizers.
		(e) and (f) show two iterations of \bo where the first point is chosen from the two points maximizing 2-\ei, and the second one is chosen by maximizing \ei (conditioned on the observation in iteration one). \vspace{-1em}
	}
	\label{fig:intuition}
\end{figure*}

Our contributions can be summarized as follows: 
(1) We propose a general framework for efficient and nonmyopic \sed with finite horizons, inspired by the close connection between optimal sequential and batch designs. 
(2) We realize the framework on two important \sed problems: Bayesian optimization and Bayesian quadrature. This represents the first nonmyopic policy proposed for \bq.  
(3) We conduct thorough experiments demonstrating that the proposed method significantly outperforms the myopic baselines and is competitive with (if not better than) state-of-the-art nonmyopic alternatives, while being much more efficient. 

\section{\binoc}
\label{sec:methods}
We will first illustrate the intuition behind \binoc and provide explicit mathematical justification.
We will then realize \binoc for two specific \sed scenarios: \bo and \bq. Throughout the rest of this work, we will make extensive use of Gaussian processes ({\gp}s): a Gaussian process defines a probability distribution over functions, where the joint distribution of the function’s values at finitely many locations is multivariate normal; for more details, see \citep{rasmussen06}.

\textbf{Intuition.}
Consider the \bo example in Fig. \ref{fig:intuition}, where we wish to maximize a one-dimensional objective function over an interval, conditioned on initial observations at the boundary. Suppose we are allowed to design two further function evaluations. The myopic \ei policy would greedily pick the middle point first, followed by a point bisecting the left half of the domain. The resulting choices completely ignore the right half, where the maximum happens to lie. 

Now consider the following alternative for designing the observations: we first construct the optimal \emph{batch} of size two ($2$-\ei). These points can be determined relatively efficiently as recursion is not required and reflect a better approximation of the remainder of the optimization than just looking one step ahead. We then pick any point from this batch and use \ei to choose the final point given the result. This policy results in well-distributed queries and better performance. We can compare these decisions with the \emph{optimal} (but expensive) policy maximizing the full lookahead expected utility ($2$-step-\ei in Fig. \ref{fig:ei_vs_2ei}): our choices are nearly perfect.  
\subsection{The Optimal Adaptive Policy} 
\label{sec:bellman_equation}
Consider a general \sed problem with a finite horizon, $T$. 
Let the design space be $\mc{X}$, response space be $\mc{Y}$; for $x\in \mc{X}, y\in \mc{Y}$ and $\data \subseteq \mc{X} \times \mc{Y}$, let $p(y \given x, \data)$ be a probabilistic model; and let $u(\data)$ be some utility function of observed data $\data$. 
Define $u(y \given x, \data) = u\bigl(\data \cup (x, y)\bigr) - u(\data)$ to represent the marginal gain in utility after observing $y$ from experiment $x$ when $\data$ has already been observed.
Let $Q_k(x \given \data)$ be the expected utility of designing experiment $x$ after observing $\data$ when there are $k$ steps remaining, assuming all later decisions are optimal. $Q_k(x \given \data)$ can be expressed in the form of a Bellman equation as follows:
\begin{multline}
	Q_k(x \given \data) = \E_{y}[u(y \given x, \data)] +  {} \\
	\E_{y}\Big[\underset{x'}{\textrm{max }} Q_{k-1}\big(x' \given \data \cup \{(x, y)\}\big)\Big],  \label{eq:bellman_equation}
\end{multline}
where the expectation is taken with respect to $p(y \given x, \data)$.
The optimal (expected-case) policy is
\begin{equation}
	x^* = \underset{x}{\textrm{argmax }} Q_{T-i}(x \given \data_{i}), \label{eq:optimal_policy}
\end{equation}
where $\data_i$ is the observed set at iteration $i$.
The optimal policy is intractable for any moderately large horizon; in general, the complexity is $\mc{O}\left(|\mc{X}|^T|\mc{S}|^T\right)$, where $\mc{S}=\{\data \given \data \subseteq \mc{X} \times \mc{Y} \}$, and in many settings $\mc{X}$ and/or $\mc{S}$ are uncountable. 
Thus, we must find some tractable approximation to proceed.
A common solution is to simply limit the horizon to some manageable value $\ell$, e.g.\ $\ell=1$ or $2$. This is called \emph{$\ell$-step lookahead,} and is computationally efficient but \emph{myopic} as we severely limit our view of the future. It does not plan ahead and can thus make suboptimal tradeoffs between exploration and exploitation. 

\subsection{Nonmyopic Approximation via the Optimal Non-Adaptive Policy}
Suppose $T$ experiments $X=\{x_1, \dots, x_T\}$ must be designed \emph{simultaneously} given current observations $\data$. The expected marginal utility of the resulting observations is
\begin{equation}
	Q(X \given \data) = \E_{Y}[u(Y\given X, \data)],  \label{eq:batch_utility}
\end{equation}
where the expectation is taken over the joint distribution of $Y=\{y_1, \dots, y_T\}$, $p(Y\given X, \data)$.
Rewriting \eqref{eq:batch_utility} by decomposing $X$ into $x_j$ and $X_{-j}$ where $X_{-j} = X\setminus \{x_j\}$, we have (by telescoping sum)
\begin{multline}
	Q(X \given \data) = \E_{y_j}[u(y_j\given x_j, \data)] + \\ 
	\E_{y_j}\Big[ Q\big(X_{-j}\given \data\cup \{(x_j, y_j)\}\big) \Big]. \label{eq:batch_utility_decomposed}
\end{multline}
Let $X^* \in \textstyle{\argmax_{X} Q(X\given \data)}$ be an optimal batch of experiments. 
For any $x^*_j \in X^*$, it follows that
\begin{multline}
	\E_{y^*_j}\Big[ Q\big(X^*_{-j}\given \data\cup \{(x^*_j, y^*_j)\}\big) \Big] = \\ \underset{X_{-j}}{\textrm{max }} \E_{y^*_j}\Big[ Q\big(X_{-j}\given \data\cup \{(x^*_j, y^*_j)\}\big) \Big],
\end{multline}
as otherwise we can construct a batch with higher utility than $Q(X^*\given \data)$.
Therefore, given that the expected reward of the entire batch can be decomposed using \eqref{eq:batch_utility_decomposed}, choosing any experiment $x^* \in X^*$ is equivalent to solving the following optimization: $x^* \in \argmax_{x} B(x \given \data)$ where 
\begin{multline}
    B(x \given \data) = \E_{y}[u(y\given x, \data)]  \\
	+ \underset{X':|X'|=T-1}{\max}  \E_{y} \Big[ Q\big(X' \given \data\cup \{(x, y)\}\big) \Big]. \label{eq:nonmyopic_approximation}
\end{multline}
Comparing \eqref{eq:nonmyopic_approximation} and the Bellman equation \eqref{eq:bellman_equation}, we see two differences:
1) the expectation and maximization are exchanged in the future utility term and 
2) the adaptive utility is replaced by a non-adaptive counterpart.
As such, \eqref{eq:nonmyopic_approximation} is clearly a \emph{lower bound of the true expected utility}:
\begin{align}
    \underset{X':|X'|=T-1}{\max} &\E_{y} \Big[ Q\big(X' \given \data\cup \{(x, y)\}\big) \Big]  \nonumber \\  
    {}\le{} 
    &\E_{y} \left[ \underset{X':|X'|=T-1}{\max} Q\big(X' \given \data\cup \{(x, y)\}\big) \right]  \nonumber \\ 
    {}\le{}
    &\E_{y}\left[\underset{x'}{\textrm{max }} Q_{T-1}\big(x' \given \data\cup \{(x, y)\}\big) \right].
\end{align}

This is illustrated in Fig. \ref{fig:ei_vs_2ei}: 2-step-\ei corresponds to \eqref{eq:bellman_equation}, and 2-\ei to \eqref{eq:nonmyopic_approximation}. 
An interesting open question is the tightness of this bound, closely related to the so-called \emph{adaptivity gap} \citep{jiang2018efficient, krause2007nonmyopic}.
The similarity between these formulations provides mathematical justification for using \eqref{eq:nonmyopic_approximation} to approximate the optimal policy. 
Note that \eqref{eq:nonmyopic_approximation} is exactly equal to \eqref{eq:bellman_equation} if the remaining experiments become conditionally independent given the observed data,
in which case there is no advantage to adaptation.

\begin{algorithm}[!h]
\caption{\binoc}
\label{alg:pseudocode}
\begin{algorithmic}
    \STATE \textbf{Input:} design space $\mc{X}$, response space $\mc{Y}$,  model $p(y \given x, \data)$, utility function $u(y\given x, \data)$, budget $T$
    \STATE \textbf{Output:} $\data$, a sequence of experiments and observations
    \FOR{$i\leftarrow 0$ to $T-1$}
    \STATE Compute the optimal batch $X^*$ of size $T-i$
    \STATE Pick an experiment $x^*\in X^*$\ and observe response $y^*$
    \STATE Augment $\data=\data\cup\{(x^*,y^*)\}$
    \ENDFOR
\end{algorithmic}
\end{algorithm}

\binoc is summarized in Algorithm \ref{alg:pseudocode}. The primary computational cost comes from computing the optimal batch, a high-dimensional optimization problem. 
For the examples considered below (\bo and \bq), this optimization can be done using gradient-based methods and we show empirically that \binoc runs much faster than previously proposed nonmyopic methods (see section \ref{sec:experiments}). 
Note that while we do use a batch method, it is only as a subroutine. Algorithm \ref{alg:pseudocode} is for \emph{sequential} experimental design: in each iteration, we only observe the outcome of one experiment. 

\section{\binoc for Bayesian Optimization}
Consider the task: $x^* =  \textstyle{ \argmax_{x\in\mc{X}} f(x); }$
in this paper, we model $f$ with a \gp.
Suppose we have a budget of $T$ function evaluations. Once the budget has been expended, we recommend the point with the highest observed value as the maximizer of $f$. 
In this setting, our goal is to \emph{sequentially} select a set $X = \{x_1,x_2,\ldots,x_T\}$ of $T$ points from $\mc{X}$ such that $\max \{ y_j \} $ is maximized, where $y_j = f(x_j)$.

Let $\data_0$ be a set of initial observations, and $y_0 = \max_{(x,y)\in \data_0} y$ is the initial best observed value. We define the utility function as the improvement over $y_0$:
\begin{equation}\label{eq:utility_function}
	u(Y \given X, \data_0) = \left( \underset{x_j\in X}{\textrm{max }} y_j - y_0\right)\mathclap{^+}\textrm{ }, 
\end{equation}
where $c^+ = \max(c, 0)$. 
Defining the utility as improvement allows us to write the expected utility as a Bellman equation with the same form as \eqref{eq:bellman_equation} (derivation in the appendix):
\begin{equation}
	EI_k(x) = EI_1(x) + \E_y\left[ \textstyle{ \max_{x'} } EI_{k-1}(x'\given x, y) \right],
\end{equation}
where $EI_k(x)$ is the expected improvement of $k$ adaptive decisions starting from $x$, and $EI_{k-1}(x' \given x,y)$ is an expectation taken over the posterior belief of $f$ after further conditioning on the observation $(x,y)$ and replacing $y_0$ by $\max(y_0, y)$.
Observe that $\argmax_x EI_1(x)$ exactly corresponds to the popular \emph{expected improvement} (\ei) policy \citep{movckus1975bayesian}, which is one-step optimal; 
$EI_2(x)$ is already analytically intractable as it requires an expensive numerical integration: the integrand is $\max_{x'} EI_1(x'\given x, y)$ and entails global optimization!

To apply \binoc, we optimize the batch \ei objective, also known as \qei, via the recently developed reparameterization trick and Monte Carlo approximation \citep{wang2016parallel}. 
Then we pick a point from the optimal batch; how to pick this point is discussed later. 
\binoc trivially extends to other utility functions such as knowledge gradient \citep{wu2016parallel}, probability of improvement \citep{kushner1964new} and predictive entropy \citep{shah2015parallel} by replacing \qei appropriately.

\section{\binoc for Bayesian Quadrature}
\label{sec:bq}
Consider a non-analytic integral of the form $	Z=\int f(x)\pi(x)\intd{x},$ where $f(x)$ is a likelihood function and $\pi(x)$ is a prior. 
Such integrals frequently occur in Bayesian inference (e.g., Bayesian model selection and averaging). 
Bayesian quadrature operates by placing a \gp on the integrand and then minimizing the posterior variance of $Z$:
\begin{equation}
	\textstyle{
		\mathrm{Var}[Z\given X] =\iint K_X(x,x')\pi(x)\pi(x')\intd{x}\intd{x'},
	}
	\label{eq:bq_utility}
\end{equation} 
where $X=\{x_1,x_2,\dots,x_T\}$ is a set of $T$ points that needs to be optimized, and $K_X(x,x')$ is the posterior covariance after conditioning on observations at $X$.
If the \gp hyperparameters are fixed, the optimal design $X^*= \textstyle{\mathrm{argmin}_X\mathrm{Var}[Z\given X]}$ can be precomputed, as the posterior covariance of a \gp does not depend on the observed values $f(X)$; this effectively eliminates the need for sequential experimental design in this setting. 

However, in general the hyperparameters are not fixed \emph{a priori}, but are instead learned iteratively in light of new observations. Furthermore, when the integrand is known to be \emph{positive} (e.g., a likelihood function), it is often a good practice to place a \gp on some non-linear transformation of $f$, such as $\sqrt{f}$ or $\log(f)$ \citep{osborne12, gunter14, chai19a}.
As a result, the posterior \gp must be approximated (e.g., by moment matching), which causes the posterior covariance to depend on the observed values. 
In these cases adaptive sampling becomes critical.

The adaptive version of $\mathrm{Var}[Z\given X]$ is computationally expensive to evaluate so \citet{gunter14} proposed the use of \emph{uncertainty sampling} (\unct) \citep{lewis94, settles10} as a surrogate, i.e.\ sequentially evaluating the location with the largest variance. This greedily minimizes the entropy of the integrand instead of the integral.

Formally, we use the differential entropy of the multivariate Gaussian $f(X)$ as the utility function:
\begin{equation}
	H(Y\given X)=\tfrac{1}{2}\log\Big(\det\big({2\pi e\,K(X,X)}\big)\Big).
	\label{eq:bq_approx}
\end{equation} 
Using the chain rule for differential entropy, this quantity can be expressed in the same form as \eqref{eq:bellman_equation}:
\begin{equation}
	H(Y\given X)=H(y_j\given x_j)+\E_{y_j}[ H(Y_{-j}|X_{-j}, x_j, y_j) ].  
	\label{eq:bq_deriv}
\end{equation} 
Note that $\textstyle{\argmax_{x_j} H(y_j\given x_j)}$ corresponds to the sequential uncertainty sampling policy.
To apply \binoc for \bq, we must find $\textstyle{\argmax_X} H(Y\given X)$, which is the mode of a determinantal point process (\acro{DPP}) \citep{kulesza12} defined over $q=|X|$ points. 
This can be done using gradient-based optimization. Note that this formulation can be applied to active learning of {\gp}s, where uncertainty sampling is also a common strategy.

\textbf{Practical Considerations.} 
\label{sec:practical}
Some practical issues arise when applying \binoc to real problems. First, given an optimal batch, how should one go about selecting a point from this batch? We considered several options: selecting the point with the highest expected immediate reward or randomly selecting a point, either proportional to their expected immediate reward or simply uniformly. 
Empirically, we found that ``best'' and ``proportional sampling'' perform similarly while ``uniform sampling'' is worse than the other two methods.
 
Second, given that \binoc is only an approximation to the optimal policy, it is not necessarily true that setting $q$ to the exact remaining budget is the best. In theory, if the model is perfect, then full lookahead is optimal. However, in practice, the model is always wrong and thus planning too far ahead could hurt the empirical performance \citep{yue2019why}. 
Further, smaller values of $q$ result in more efficient computation. We empirically study the choice of $q$ in section \ref{sec:experiments}.

\section{Related Work}
\label{sec:related}
General introductions to approximate dynamic programming (\acro{DP}) can be found in \citet{bertsekas2017dynamic, powell_2011}.
On the subject of nonmyopic \bo, \citet{osborne2009gaussian} derived the optimal policy for \bo, and demonstrate that it is possible to approximately compute (with great effort) the two-step lookahead policy for low-dimensional functions and that it generally performs better than the one-step policy.
\citet{ginsbourger2010towards} also derived the optimal policy and gave an explicit example where two-step \ei is better than one-step \ei in expectation with a desired degree of statistical significance. 
\citet{gonzalez2016glasses} proposed a nonmyopic approximation of the optimal policy, known as \glasses, by simulating future decisions using a batch \bo method.
\footnote{The name \binoc is inspired by \glasses.}
\citet{jiang2017efficient, jiang2018efficient} proposed a nonmyopic policy for (batch) active search, which can be understood as a special case of \bo with cumulative reward, using a similar idea.
\citet{lam2016bayesian} proposed to use rollout for \bo, which is a classic approximate \acro{DP} method  \citep{bertsekas2017dynamic}. 
\citet{yue2019why} presented theoretical justification for rollout, and gave theoretical and practical guidance on how to choose the rollout horizon. 
\citet{ling2016gaussian} proposed a branch-and-bound near-optimal policy for \gp planning assuming that the reward function is Lipschitz continuous, and applied it to \bo and active learning.
\citet{wu2019practical} proposed a gradient-based optimization of two-step \ei, but each evaluation of two-step \ei still requires a quadrature subroutine with an expensive integrand: optimization of one-step \ei. 

Of these, \glasses and rollout are most related to \binoc.
\glasses's acquisition function shares almost the same form as \eqref{eq:nonmyopic_approximation}, except the future batch $X'$ is constructed using a heuristic batch policy, instead of optimized with the \qei objective. The batch policy adds points one by one by optimizing the sequential \ei function penalized at locations already added to the batch \citep{gonzalez2016batch}, and the expected utility of the chosen batch is estimated using expectation propagation.

\begin{table*}[!h]
    \small
	\centering
	\caption{Average \gap over 100 repeats on ``hard'' synthetic functions.} 
	
\begin{tabular}{lllllllllllll}
\toprule
&{Rand} & {EI} & {2.EI.b} & {2.EI.s} & {3.EI.b} & {3.EI.s} & {4.EI.b} & {4.EI.s} & {10.EI.b} & {10.EI.s} & {12.EI.s} & {15.EI.s}\\\hline
{eggholder} & 0.498   &  0.613   &  0.614   &  0.633   &  0.604   &  0.657   &  0.646   &  \textit{\textcolor{blue}{0.694  }} &  0.622   &  \textit{\textcolor{blue}{0.704  }} &  \textbf{0.738  } &  \textit{\textcolor{blue}{0.694  }} \\  
{dropwave} & 0.486   &  0.439   &  0.507   &  0.531   &  0.473   &  \textit{\textcolor{blue}{0.552  }} &  0.467   &  0.514   &  0.397   &  \textit{\textcolor{blue}{0.591  }} &  \textbf{0.598  } &  \textit{\textcolor{blue}{0.585  }} \\  
{shubert} & 0.355   &  0.408   &  0.366   &  \textit{\textcolor{blue}{0.441  }} &  \textit{\textcolor{blue}{0.394  }} &  \textbf{0.507  } &  0.388   &  \textit{\textcolor{blue}{0.484  }} &  0.305   &  \textit{\textcolor{blue}{0.455  }} &  \textit{\textcolor{blue}{0.479  }} &  \textit{\textcolor{blue}{0.465  }} \\  
{rastrigin4} & 0.374   &  0.801   &  0.769   &  0.775   &  0.817   &  \textit{\textcolor{blue}{0.821  }} &  \textbf{0.840  } &  0.805   &  0.797   &  0.804   &  0.793   &  0.799   \\  
{ackley2} & 0.358   &  0.821   &  0.825   &  0.823   &  0.819   &  \textit{\textcolor{blue}{0.869  }} &  0.812   &  \textit{\textcolor{blue}{0.872  }} &  0.801   &  \textbf{0.892  } &  \textit{\textcolor{blue}{0.886  }} &  \textit{\textcolor{blue}{0.888  }} \\  
{ackley5} & 0.145   &  0.509   &  0.544   &  0.509   &  0.601   &  0.550   &  0.596   &  0.592   &  \textbf{0.636  } &  0.606   &  \textit{\textcolor{blue}{0.627  }} &  \textit{\textcolor{blue}{0.626  }} \\  
{bukin} & 0.600   &  \textit{\textcolor{blue}{0.849  }} &  0.856   &  0.855   &  \textit{\textcolor{blue}{0.872  }} &  \textit{\textcolor{blue}{0.859  }} &  0.864   &  \textit{\textcolor{blue}{0.865  }} &  \textbf{0.878  } &  0.850   &  0.829   &  \textit{\textcolor{blue}{0.853  }} \\  
{shekel5} & 0.038   &  0.286   &  0.311   &  0.320   &  0.330   &  0.343   &  0.342   &  0.344   &  \textit{\textcolor{blue}{0.374  }} &  \textit{\textcolor{blue}{0.373  }} &  \textit{\textcolor{blue}{0.358  }} &  \textbf{0.395  } \\  
{shekel7} & 0.045   &  0.268   &  0.346   &  0.313   &  0.349   &  0.325   &  0.352   &  0.370   &  \textit{\textcolor{blue}{0.399  }} &  0.358   &  \textbf{0.412  } &  \textit{\textcolor{blue}{0.386  }} \\  \hline
{Average} & 0.322   &  0.555   &  0.571   &  0.578   &  0.584   &  0.609   &  0.590   &  0.616   &  0.579   &  \textit{\textcolor{blue}{0.626  }} &  \textbf{0.635  } &  \textit{\textcolor{blue}{0.632  }} \\  
\bottomrule
\end{tabular}

	\label{table:bo_synthetic}
\end{table*}

Rolling out two steps using \ei as the heuristic policy is exactly equivalent to the two-step lookahead policy, up to quadrature error.
Mathematically, the rollout acquisition function can also be written in a similar form as \eqref{eq:nonmyopic_approximation}, except $X'$ is adaptively constructed, depending on sampled values of $y$ instead of globally (irrespective of $y$) constructed or optimized as in \glasses and \binoc. Both rollout and \glasses are very expensive to compute.

While we are unaware of any existing work on nonmyopic \bq, there has been some prior work on nonmyopic active learning of {\gp}s.
\citet{krause2007nonmyopic} derived the adaptivity gap for active learning of {\gp}s under two utility functions. They also proposed a nonmyopic method for active learning of {\gp}s which separates the process into an exploration phase and an exploitation phase. They considered different acquisition functions for the exploration phase; notably, the implicit exploration (\acro{IE}) method is comparable to the uncertainty sampling baseline in subsection \ref{sec:bq_results}.
\citet{hoang14} developed a method for active learning of {\gp}s that does away with separate exploration and exploitation phases and instead naturally trades off between the two. Their proposed policy, $\varepsilon$-\acro{BAL}, approximates the solution to the Bellman formulation of the active {\gp} learning problem using a truncated sampling method. They analyzed the theoretical performance of their method and also developed a pruning-based anytime version of their method.

The setting of our \bq work (integration of non-negative integrands) and active learning of {\gp}s appear related yet are fundamentally different. The cited works focus exclusively on learning the hyperparameters of the \gp. In our setting, the use of a transformation to model non-negativity introduces adaptivity beyond the \gp hyperparameters: even if the true \gp hyperparameters are known \emph{a priori}, the nonlinear transformation causes the approximate \gp posterior to depend on the observed values.

\section{Experiments}
\label{sec:experiments}
We designed our experiments to broadly test the performance and computational cost of \binoc relative to notable myopic and nonmyopic baselines for \bo and \bq. 
We also conducted a thorough exploration of the \binoc design choices: the number of steps to look ahead and how to select a point from the optimal batch. 

The primary takeaways of our experimental results are that \binoc outperforms myopic baselines while running only slightly slower and is at least as good as previously proposed nonmyopic methods while running orders of magnitude faster.
This places it \emph{on the Pareto front} of the running time--performance tradeoff in policy design.
Furthermore, \binoc clearly demonstrates \emph{distinctively nonmyopic behavior} on both \bo and \bq tasks, two entirely different \sed problems. 

We use the following nomenclature to describe \binoc: 
our nonmyopic \bo method will be denoted as ``$q$.\acro{EI}.s'' or ``$q$.\acro{EI}.b'', where $q$ is the batch size and ``s'' represents sampling from the batch while ``b'' means choosing the ``best.''
For \bq, we replace ``\acro{EI}'' with ``\acro{DPP}.'' 
In addition to the myopic methods, \ei and \unct, we also compare against rollout for both tasks and \glasses for \bo.
\footnote{We did not compare against a \bq-equivalent of \glasses as no such method has been published.}
Each rollout method is denoted as ``$q$.\R.$n$'', where $q$ represents the number of steps to roll out, and $n$ is the number of samples used to estimate the expectations encountered in each step. 
Each \glasses method is denoted as ``$q$.\acro{G}'' where $q$ represents the size of the simulated batch. 
We use \acro{DIRECT} \citep{Jones2009} to optimize the \glasses and rollout acquisition functions, following \citet{gonzalez2016glasses}. 
For all nonmyopic methods, when the remaining budget $r < q$, we set $q = r$. 
Thus the final decision is always made (optimally) with one-step lookahead.

For all experiments, we start with $2d$ randomly-sampled observations and perform $20d$ further iterations, where $d$ is the function's dimensionality. Unless otherwise noted, all results presented are aggregated over 100 repeats with different random initializations. For all tabulated results, the best method is indicated in bold and the entries not significantly worse than the best (under a one-sided paired Wilcoxon signed-rank test with $\alpha=0.05$) are in blue italics. 
\subsection{BO Results}
\label{sec:bo_results}
We implemented our nonmyopic \bo policy and all baselines using \texttt{BoTorch},\footnote{\url{https://github.com/pytorch/botorch}} which contains efficient \ei and \qei implementations.
We present experiments for two rollout variants: ``2.\R.10'' and ``3.\R.3.'' As we will see, rolling out with horizon two is already very expensive even for just ten $y$ samples. Gauss--Hermite quadrature is used for rollout as in \citet{lam2016bayesian}. 
We also present experiments for two \glasses variants: ``2.\acro{G}'' and ``3.\acro{G}''.
\footnote{With help from the authors of \citep{gonzalez2016glasses}, we implemented an advanced version of \glasses in \texttt{BoTorch}, using quasi Monte Carlo instead of expectation propagation to estimate the expected improvement of the batch, a standard practice for computing q\EI in state of the art \bo packages such as \texttt{BoTorch}.}

We use {\gp}s with a constant mean and a Mat{\'e}rn $\nicefrac{5}{2}$ \acro{ARD} kernel to model the objective function, the default in \texttt{BoTorch}. 
We tune hyperparameters every iteration by maximizing the marginal likelihood using \acro{L-BFGS-B}. We also maximize the \qei acquisition function with \acro{L-BFGS-B}. Complete details can be found in our attached code. We use the \gap measure to evaluate the performance: \gap $={(y_i - y_0)}\,/\,{(y^*-y_0)}$, where $y_i$'s are maximum observed values and $y^*$ is the true optimal value; we convert all problems to maximization problems by negating if necessary.

\textbf{Synthetic Functions}.
In this section, we focus on demonstrating the superior performance of our method over \ei on nine ``hard'' benchmark functions. 
These nine functions are selected by first running experiments on 31 functions\footnote{\url{https://www.sfu.ca/~ssurjano/optimization.html}} with 30 repeats (see Table \ref{table:all_synthetic_functions} in appendix). 
We then select the ones where \ei terminates with average \gap < 0.9.
We believe nonmyopic methods are more advantageous on challenging functions; by first identifying these hard problems, we will gain more insight into the various policies.
To put the \bo performance into perspective, we also include a comparison against a random baseline, ``Rand.''

Table \ref{table:bo_synthetic} shows the average \gap at termination. We summarize the results as follows:
(1) All $q$.\ei.s variants perform significantly better than \ei on average, with 12.\ei.s being the best and outperforming \ei by a large margin.
(2) The $q$.\ei.s variants are consistently better than the $q$.\ei.b variants (for better spacing we did not show results for 12.\ei.b and 15.\ei.b).
(3) The performance of our method generally improves as we increase $q$, up to 12.

\begin{figure}[t!]
		\centering
 		\includegraphics[width=\columnwidth]{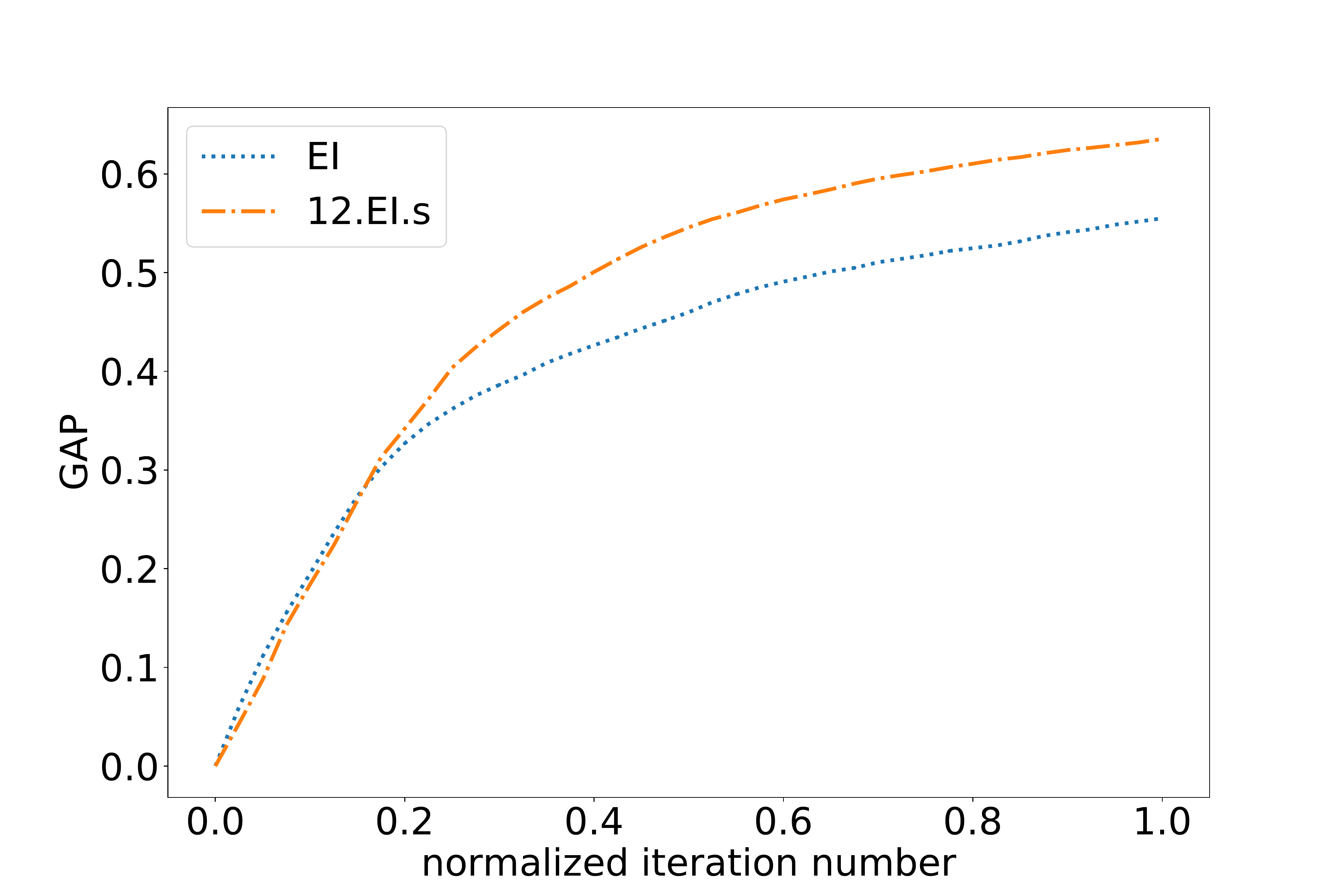}
		\caption{Average \gap over nine synthetic functions demonstrating the nonmyopic behavior of 12.\ei.s.}
		\label{fig:synthetic_iter_aggregated_nonmyopic_behavior}
\end{figure}

\begin{table*}[!h]
    \small
	\centering
	\caption{Average \gap over 50 repeats on real functions.}
	\begin{tabular}{lllllllllll}
\toprule
&{EI} & {2.EI.s} & {3.EI.s} & {4.EI.s} & {6.EI.s} & {8.EI.s} & {2.G} & {3.G} & {2.R.10} & {3.R.3}\\\hline
{\svm} & 0.738   &  \textit{\textcolor{blue}{0.913  }} &  \textbf{0.940  } &  \textit{\textcolor{blue}{0.911  }} &  \textit{\textcolor{blue}{0.937  }} &  0.834   &  \textit{\textcolor{blue}{0.881  }} &  0.898   &  \textit{\textcolor{blue}{0.930  }} &  \textit{\textcolor{blue}{0.928  }} \\  
{\lda} & 0.956   &  \textbf{1.000  } &  \textit{\textcolor{blue}{0.996  }} &  \textit{\textcolor{blue}{0.993  }} &  0.982   &  \textit{\textcolor{blue}{0.995  }} &  \textit{\textcolor{blue}{1.000  }} &  \textit{\textcolor{blue}{0.999  }} &  \textit{\textcolor{blue}{0.999  }} &  \textit{\textcolor{blue}{1.000  }} \\  
{LogReg} & 0.963   &  \textit{\textcolor{blue}{0.998  }} &  \textit{\textcolor{blue}{1.000  }} &  \textit{\textcolor{blue}{0.999  }} &  \textit{\textcolor{blue}{0.999  }} &  \textbf{1.000  } &  0.989   &  0.911   &  0.965   &  0.948   \\  
{NN Boston} & \textit{\textcolor{blue}{0.470  }} &  \textit{\textcolor{blue}{0.467  }} &  \textit{\textcolor{blue}{0.478  }} &  \textit{\textcolor{blue}{0.460  }} &  \textit{\textcolor{blue}{0.502  }} &  \textit{\textcolor{blue}{0.467  }} &  0.455   &  \textbf{0.512  } &  \textit{\textcolor{blue}{0.503  }} &  \textit{\textcolor{blue}{0.482  }} \\  
{NN Cancer} & 0.665   &  0.627   &  0.654   &  0.686   &  0.700   &  0.686   &  \textbf{0.806  } &  \textit{\textcolor{blue}{0.755  }} &  0.708   &  0.698   \\  
{Robot pushing 3d} & 0.928   &  \textit{\textcolor{blue}{0.960  }} &  \textit{\textcolor{blue}{0.962  }} &  \textit{\textcolor{blue}{0.957  }} &  \textbf{0.962  } &  \textit{\textcolor{blue}{0.961  }} &  \textit{\textcolor{blue}{0.955  }} &  0.951   &  \textit{\textcolor{blue}{0.955  }} &  \textit{\textcolor{blue}{0.954  }} \\  
{Robot pushing 4d} & \textit{\textcolor{blue}{0.730  }} &  0.726   &  0.695   &  0.695   &  0.736   &  0.697   &  \textit{\textcolor{blue}{0.765  }} &  \textbf{0.786  } &  \textit{\textcolor{blue}{0.770  }} &  \textit{\textcolor{blue}{0.745  }} \\ \hline
{Average} & 0.779   &  \textit{\textcolor{blue}{0.813  }} &  \textit{\textcolor{blue}{0.818  }} &  0.815   &  \textit{\textcolor{blue}{0.831  }} &  0.806   &  \textbf{0.836  } &  \textit{\textcolor{blue}{0.830  }} &  \textit{\textcolor{blue}{0.833  }} &  \textit{\textcolor{blue}{0.822  }} \\  
\bottomrule
\end{tabular}

	\label{table:bo_real}
\end{table*}

Perhaps more interestingly, we can clearly observe the nonmyopic behavior of 12.\ei.s as shown in Figure \ref{fig:synthetic_iter_aggregated_nonmyopic_behavior}:
it is initially outperformed by the myopic \ei as it explores the space. However, our method catches up to \ei at $\sim$20\% of the budget (on average) as it transitions to exploiting its findings until finally, it outperforms \ei by a large margin. 
This behavior indicates that our method seamlessly navigates the exploration/exploitation tradeoff without the need for any external intervention.

\textbf{Real World Functions.}
In this section, we compare our method against popular nonmyopic baselines: rollout and \glasses.
We present results on hyperparameter tuning functions used by \citet{snoek2012practical, wangICML2017b,malkomes2018automating}.
These functions are evaluated on a predefined grid, so we first compute all policies (except \ei) using continuous optimization, then pick the closest point from the grid.

Table \ref{table:bo_real} shows the results averaged over 50 repeats.
We only show the ``sampling'' variants of our method; full results can be found in Table \ref{table:real_all_qei_variants} 
First we see again all $q$.\ei.s variants outperform \ei by a large margin, with $q=6$ achieving the best results. 
Comparing 6.\ei.s with the nonmyopic baselines, 2.\acro{G} is the best, but the difference of 0.005 is negligible; 
the $p$-value under a one-sided paired signed-rank test for 6.\ei.s against 2.\acro{G} is $0.4257$.

\begin{figure}[h!]
		\centering
		\caption{mean \gap with error bars at termination versus time per iteration (in log scale) averaged over the seven real functions.}
		\includegraphics[width=\columnwidth]{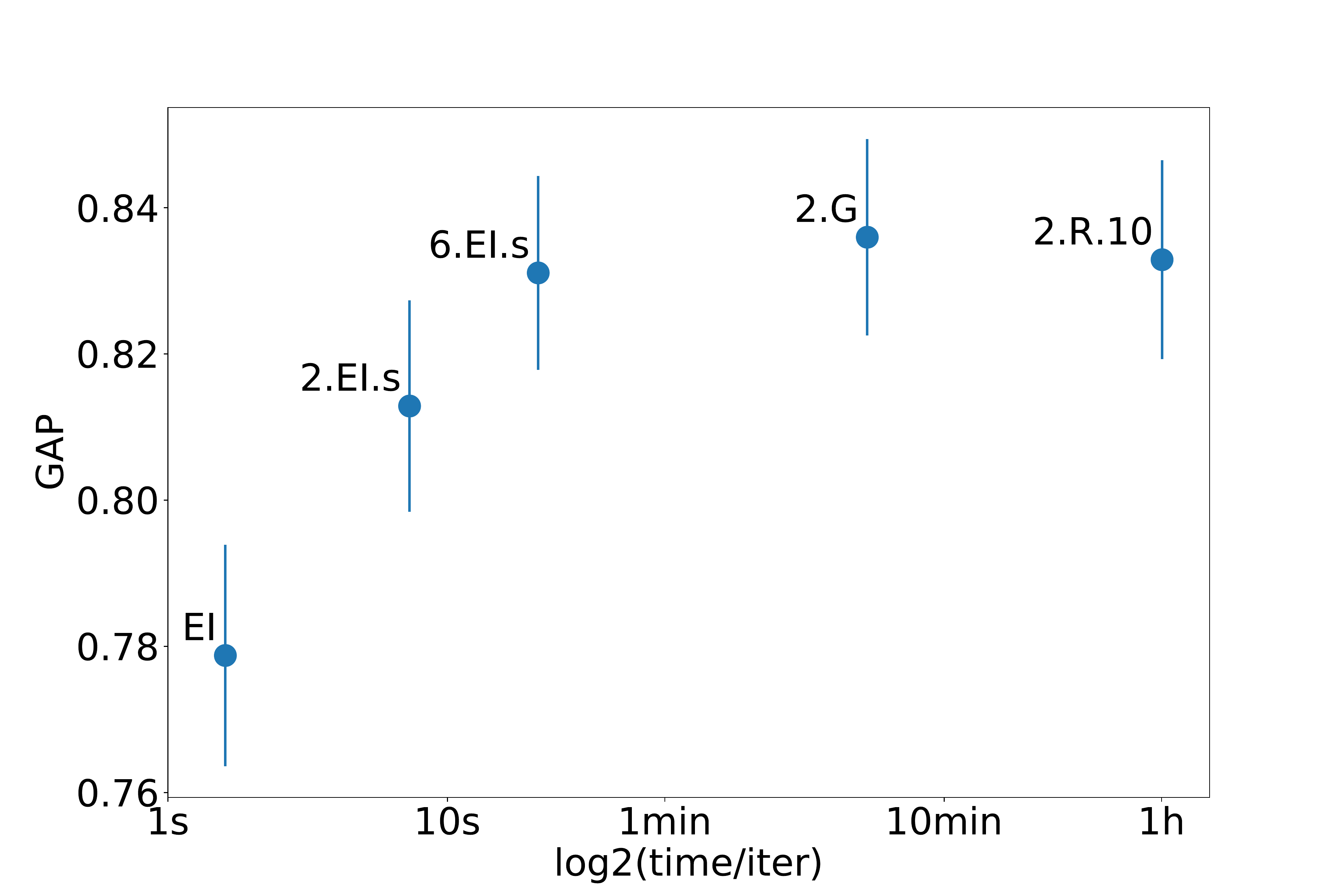}
		\label{fig:real_time_aggregated}
\end{figure}

We now focus on comparing the time cost of the tested methods. 
Figure \ref{fig:real_time_aggregated} shows the average \gap versus average time per iteration; the average is taken over 350 experiments (seven functions with 50 repeats each); error bars are also plotted. 
We again see that our methods are not significantly different from rollout and \glasses in terms of \gap performance, but are considerably faster in terms of average time cost per iteration (note the log scale on the time axis). 
Clearly, our method lies on the Pareto front in terms of computational cost and performance.  

We also attempted to compare with the recently published practical two-step \ei method \citep{wu2019practical}, which is intended to be a more efficient version of our 2.\R.$n$. As far as we understand, the difference is first- versus zeroth-order optimization of the acquisition function. 
In fact, our implementation of rollout already supports gradient-based optimization thanks to automatic differentiation. However, we did not find it considerably faster than using \acro{DIRECT}. 
We leave it to future work to optimize the implementation and compare with our method. 

It is also possible to further improve rollout's performance using an adaptive rolling horizon in light of a recent study \citep{yue2019why}, but it would be even more expensive to compute. 
In fact, Figure 1 in \citep{yue2019why} shows that with their adaptive horizon approach, the most frequently chosen horizon was two. 

\begin{table*}[!h]
    \small
	\centering
	\caption{Median fractional error values over 100 repeats on all \bq functions.}
	
\begin{tabular}{llllllllllll}
\toprule
& {UNCT} & {2.DPP.b} & {3.DPP.b} & {10.DPP.b} & {2.DPP.s} & {3.DPP.s} &  {10.DPP.s} & {2.R.10} & {3.R.3} \\ \hline
{cont} & 0.045 & 0.052 & 0.055 & 0.059 & 0.039 & 0.037 & \textbf{0.029} & 0.036 & 0.045 \\  
{corner} & 0.265 & 0.206 & 0.137 & 0.065 & \textbf{0.047} & 0.078 & 0.132 & 0.074 & 0.063 \\  
{discont} & \textit{\textcolor{blue}{0.523}} & \textit{\textcolor{blue}{0.511}} & \textit{\textcolor{blue}{0.488}} &  \textbf{0.446} & 0.572 & 0.610 & 0.590 & 0.537 & 0.577 \\  
{Gauss} & \textit{\textcolor{blue}{0.004}} & 0.004 & 0.005 & 0.006 &  \textbf{0.003} & \textit{\textcolor{blue}{0.003}} &  \textit{\textcolor{blue}{0.003}} & 0.004 & \textit{\textcolor{blue}{0.003}} \\
{\acro{mm}} & 0.254 & 0.207 & 0.203 & 0.207 & 0.221 & 0.161 & 0.177 &  \textit{\textcolor{blue}{0.110}} & \textbf{0.086} \\  
{prod} & 0.007 & 0.007 & \textit{\textcolor{blue}{0.007}} & 0.007 & 0.007 &  \textbf{0.006} & \textit{\textcolor{blue}{0.006}} & 0.012 & 0.012 \\ \hline 
{\acro{GP}} & 0.231 & 0.082 & \textbf{0.057} & 0.077 &  \textit{\textcolor{blue}{0.069}} & \textit{\textcolor{blue}{0.073}} & 0.116   & 0.283 & 0.248 \\  
{\acro{DLA}} & 0.019 & \textit{\textcolor{blue}{0.013}} & 0.025 &  \textit{\textcolor{blue}{0.013}} & \textit{\textcolor{blue}{0.016}} &  \textit{\textcolor{blue}{0.016}} & 0.033 & \textit{\textcolor{blue}{0.019}} & \textbf{0.011} \\ \hline 
{Average} & 0.068 & 0.056 & 0.055 & \textit{\textcolor{blue}{0.041}} &  \textbf{0.037} & \textit{\textcolor{blue}{0.043}} & 0.055 & 0.049 & 0.051 \\  
\bottomrule
\end{tabular}

	\label{table:bq_all}
\end{table*}

\begin{figure*}[!h]
	\centering
	\begin{subfigure}[b]{.65\columnwidth}
		\centering
		\includegraphics[width=\columnwidth]{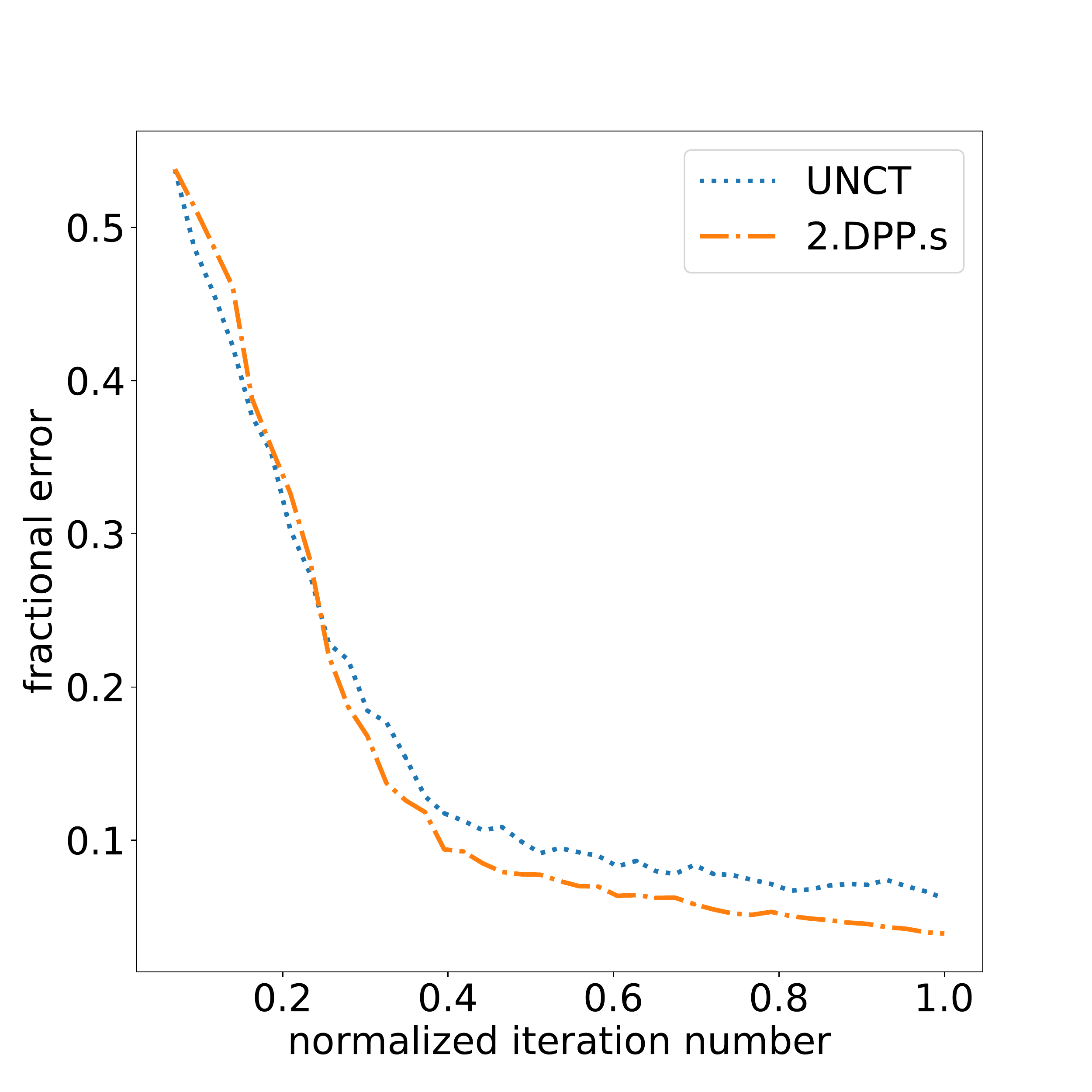}
		\subcaption{}
		\label{fig:bq_synth_its}
	\end{subfigure}
	\begin{subfigure}[b]{.65\columnwidth}
		\centering
		\includegraphics[width=\columnwidth]{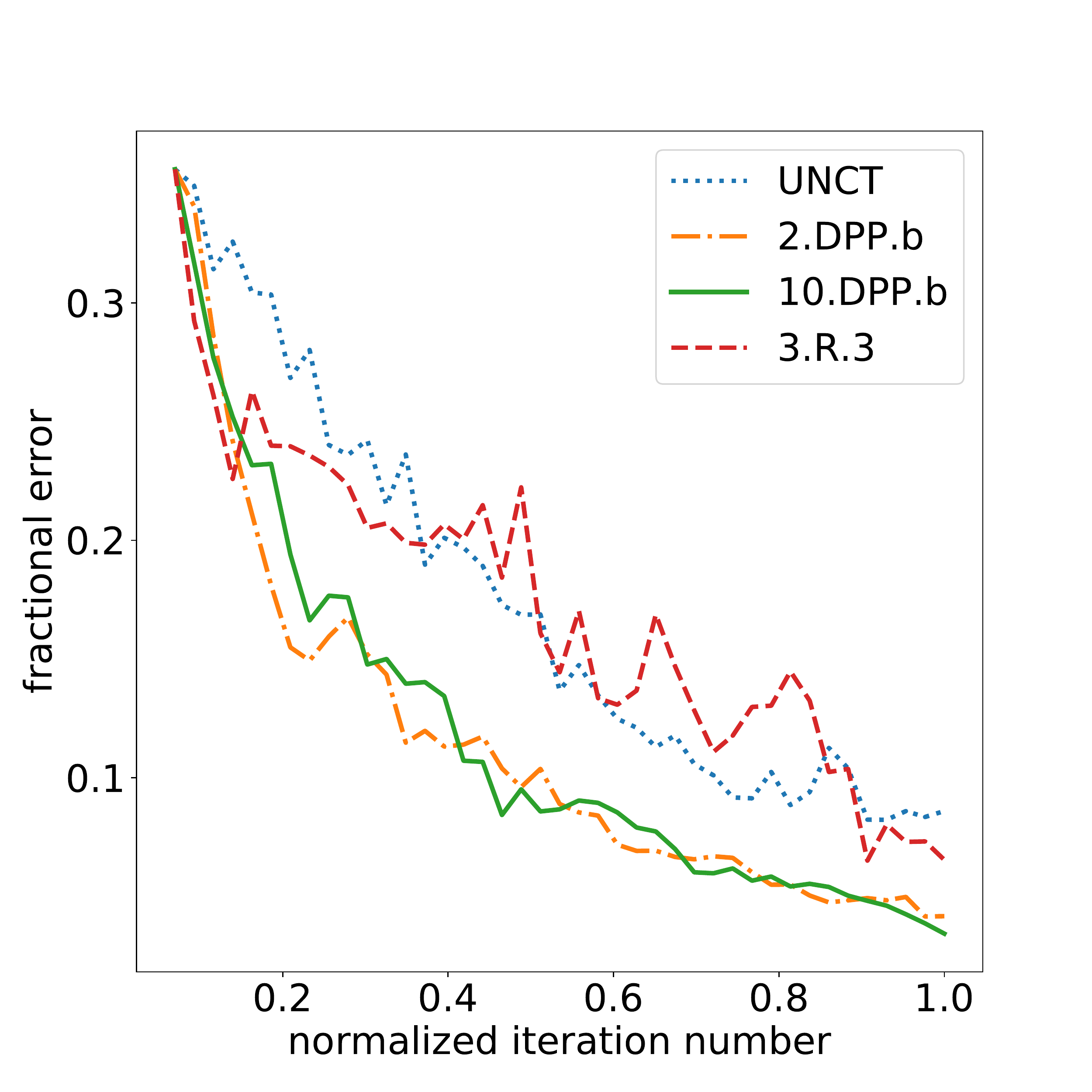}
		\subcaption{}
		\label{fig:bq_real_its}
	\end{subfigure}
	\begin{subfigure}[b]{.65\columnwidth}
		\centering
		\includegraphics[width=\columnwidth]{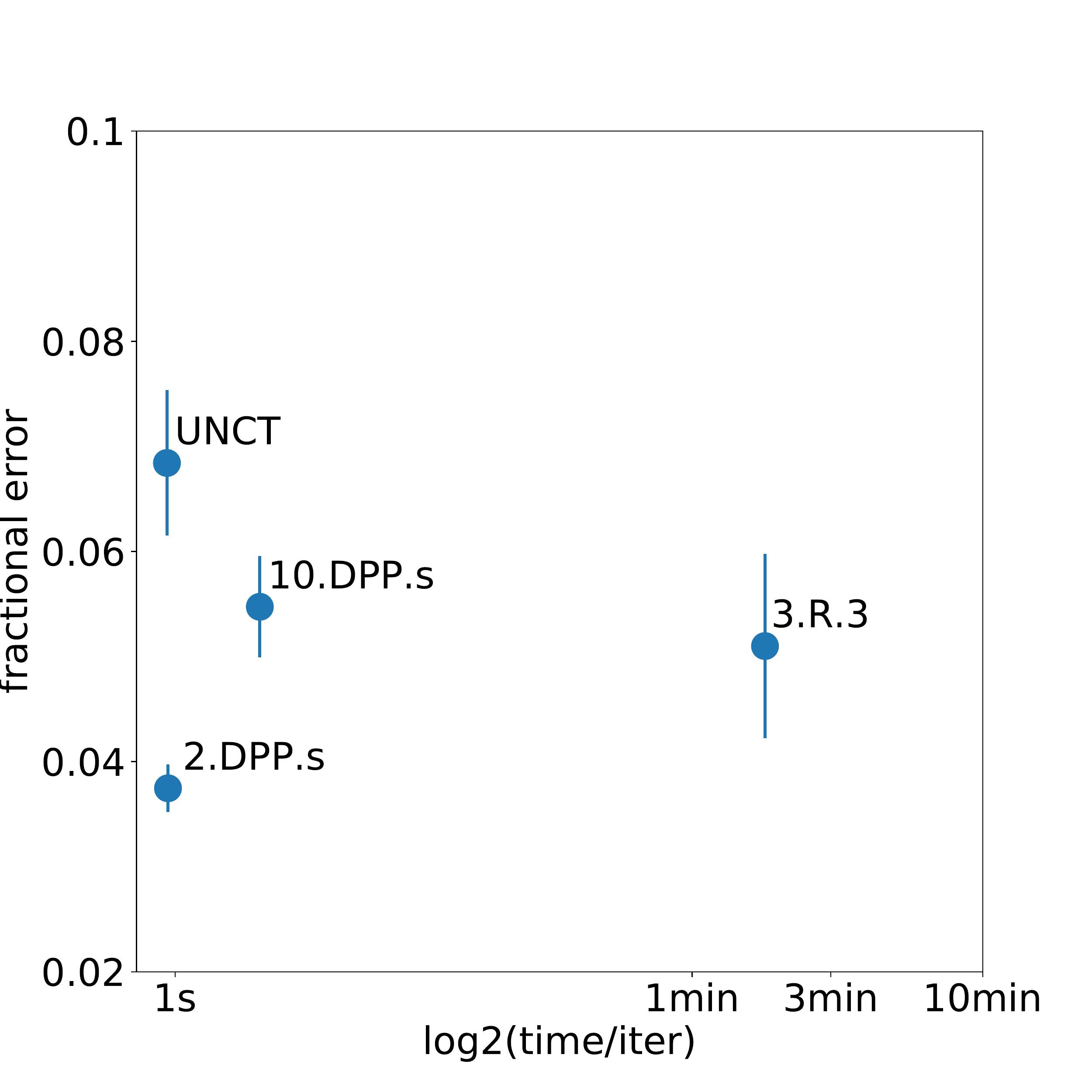}
		\subcaption{}
		\label{fig:bq_all_times}
	\end{subfigure}
	\caption{Median fractional error over 100 repeats against iterations or time. 
	(a) synthetic functions,
	(b) real functions,
	(c) all functions}
	\label{fig:bq_plots}
\end{figure*}

\subsection{BQ Results}
\label{sec:bq_results}
We implemented our nonmyopic \bq policy and all baselines 
using the \acro{GPML MATLAB} package.\footnote{\url{http://gaussianprocess.org/gpml/code/matlab}} 
For all \bq experiments, we use the framework of \citet{chai19a}: we place \gp priors on the log of the integrands as they are all non-negative. 
We use {\gp}s with a constant mean and a Mat{\'e}rn $\nicefrac{3}{2}$ \acro{ARD} kernel to model the integrands. 
We tune the \gp hyperparameters after each observation by maximizing the marginal likelihood using \acro{L-BFGS-B}. We also use \acro{L-BFGS-B} to maximize the \acro{DPP} likelihood. Complete details of our implementation can be found in our attached code.

We perform experiments on five standard benchmark synthetic functions\footnote{\url{https://www.sfu.ca/~ssurjano/integration.html}} as well as one additional synthetic benchmark and two real model likelihood functions used by \citet{chai19b}. 
The additional synthetic benchmark is: $f(x)=\prod_{i=1}^d\frac{\sin(x_i)+\nicefrac{\cos(3x_i))^2}{2}}{\nicefrac{x_i^2}{4}+0.3};$ this function was included because of its multi-modal (\acro{MM}) nature.  
We evaluate the performance of all methods using their fractional error: $\lvert Z - \hat{Z}\rvert / {Z}$ where $\hat{Z}$ is the estimate of the integral. 

Figure \ref{fig:bq_synth_its} indicates that 2.\DPP.s exhibits the same nonmyopic behavior as 12.\ei.s: it initially lags behind but eventually overtakes the myopic \unct, again suggesting a superior and automatic tradeoff of exploration and exploitation.

Figure \ref{fig:bq_all_times} shows the median fractional error versus average time per iteration; the average is taken over all 800 \bq experiments.
Our method, $2$.\DPP.s, significantly outperforms all other tested methods, is considerably faster than rollout and only slightly slower than \unct. 

Table \ref{table:bq_all} shows the median fractional error at termination for all \bq experiments. 
Fig. \ref{fig:bq_plots} shows the convergence of the fractional error as a function of both iterations and time (in $\log$ scale). 
These results corroborate many of the findings from our \bo experiments:
(1) All nonmyopic methods outperform \unct on average. 
(2) Our proposed nonmyopic methods are competitive with, if not better than, rollout while running orders of magnitude faster.

We also note that in general, $q$.\DPP.s variants tend to outperform $q$.\DPP.b variants and increasing the batch size $q$ does not consistently improve the performance. 

The primary conclusion here is the same as for \bo: \binoc significantly and consistently outperforms myopic policies while only slightly increasing  computational cost.

\section{Conclusion and Future Work}
We proposed \binoc: an efficient, nonmyopic approximation framework for finite-horizon sequential experimental design. \binoc computes an optimal batch, then picks a point from the batch. 
We gave an intuitive understanding and a mathematical justification for why this is a good approximation. 
We applied \binoc to Bayesian optimization and Bayesian quadrature, two entirely different problems, and empirically demonstrated that it significantly outperforms commonly used myopic policies while being much more efficient than popular nonmyopic alternatives.

As suggested by \citet{yue2019why}, it would be useful to derive theories to guide the choice of lookahead horizon $q$ for our method. 
Another interesting theoretical question is whether we can provide explicit bounds for the adaptivity gap for a general class of problems.

\section*{Acknowledgement}
Thanks to Eytan Bakshy and Maximilian Balandat for helping with using the \texttt{BoTorch} package before its release.  

\clearpage
\bibliography{main}
\bibliographystyle{plainnat}

\newpage
\appendix

\section{Full-Lookahead Expected Improvement as Bellman Equation }
When $T=1$, i.e., there is only one evaluation left, the optimal policy degenerates to the simplest case known as \emph{expected improvement} (\ei):
\begin{equation}
	x^* = \argmax_x EI_1(x) \equiv \E[ (f(x) - y_0)^+ ].
\end{equation}

Now consider $T=2$. 
Starting from location $x$, the improvement of the next two evaluations depends on three random variables: $y\equiv f(x)$, the next evaluation location $x'$, and its value $y'\equiv f(x')$; computing the expected utility of starting from $x$ requires integrating all three variables out:
\begin{align}
	EI_2(x) = \int_{y, x', y'} &\left(\max\{y_1, y_2\} - y_0 \right)^+ p(y\given x) \nonumber \\  
						  	  p&(x'\given x, y) p(y'\given x', y, x) dy dx' dy'.
\end{align}

Given
$$
\left( \max\{y, y'\} - y_0 \right)^+ 
	= (y - y_0)^+  + \left(y' - \max(y_0, y) \right)^+
$$ 
\citep{ginsbourger2010kriging}, we have 
\begin{align}
	EI_2(x) &= \int_y (y - y_0)^+ dy \nonumber \\
			&+ \int_y \int_{x'} \int_{y'} \left(y' - \max(y_0, y) \right)^+ p(y'\given x', y, x) dy' \nonumber \\ 
			&\hphantom{+ \int_y \int_{x'} \int_{y'}} p(x'\given x, y) dx' p(y\given x) dy  \nonumber \\
			&= EI_1(x) \nonumber \\ 
			&+ \int_y \int_{x'} EI_1(x'\given x, y) p(x'\given x, y) dx' p(y\given x) dy
\end{align}

By Bellman's principle of optimality, we have 
\begin{align}
	p(x'\given x, y) = \textstyle{\delta(x' - \argmax_{x^*} EI_1(x^*\given x, y))}.
\end{align}

Therefore, 
\begin{equation}
    \int_{x'} EI_1(x'\given x, y) p(x'\given x, y) dx' = \max_{x'} EI_1(x'\given x, y),
\end{equation}
 
and hence
\begin{align}
	EI_2(x) &= EI_1(x) + \E[ \max_{x'} EI_1(x'\given x, y) ].
\end{align}

In general, we have the following Bellman equation for $k$-step expected utility
\begin{align}
	EI_k(x) &= EI_1(x) + \E[ \max_{x'} EI_{k-1}(x'\given x, y) ].
\end{align}

\setlength\tabcolsep{5pt}
\begin{table*}
	\small
	\centering
	\caption{Average \gap of 30 repeats on all 31 synthetic functions.}
	
\begin{tabular}{llllllllllll}
\toprule
&{EI} & {2.EI.b} & {2.EI.s} & {3.EI.b} & {3.EI.s} & {4.EI.b} & {4.EI.s} & {5.EI.b} & {5.EI.s} & {10.EI.b} & {10.EI.s}\\\hline
{branin} & \textit{\textcolor{blue}{1.000  }} &  1.000   &  0.999   &  1.000   &  0.999   &  \textit{\textcolor{blue}{1.000  }} &  1.000   &  \textit{\textcolor{blue}{1.000  }} &  1.000   &  \textbf{1.000  } &  0.999   \\  
{rosenbrock2} & \textit{\textcolor{blue}{0.989  }} &  \textit{\textcolor{blue}{0.978  }} &  \textit{\textcolor{blue}{0.985  }} &  \textit{\textcolor{blue}{0.990  }} &  \textit{\textcolor{blue}{0.981  }} &  0.971   &  \textit{\textcolor{blue}{0.979  }} &  \textit{\textcolor{blue}{0.969  }} &  \textbf{0.996  } &  \textit{\textcolor{blue}{0.981  }} &  0.973   \\  
{rosenbrock4} & \textit{\textcolor{blue}{0.989  }} &  \textit{\textcolor{blue}{0.989  }} &  0.988   &  0.990   &  \textit{\textcolor{blue}{0.990  }} &  \textit{\textcolor{blue}{0.991  }} &  \textit{\textcolor{blue}{0.990  }} &  \textbf{0.992  } &  0.988   &  \textit{\textcolor{blue}{0.991  }} &  \textit{\textcolor{blue}{0.989  }} \\  
{rosenbrock6} & \textit{\textcolor{blue}{0.989  }} &  0.989   &  \textit{\textcolor{blue}{0.990  }} &  \textbf{0.992  } &  \textit{\textcolor{blue}{0.990  }} &  \textit{\textcolor{blue}{0.990  }} &  \textit{\textcolor{blue}{0.990  }} &  \textit{\textcolor{blue}{0.991  }} &  \textit{\textcolor{blue}{0.990  }} &  \textit{\textcolor{blue}{0.991  }} &  0.985   \\  
{hartmann3} & 1.000   &  \textit{\textcolor{blue}{1.000  }} &  \textit{\textcolor{blue}{1.000  }} &  1.000   &  \textit{\textcolor{blue}{1.000  }} &  \textbf{1.000  } &  \textit{\textcolor{blue}{1.000  }} &  \textit{\textcolor{blue}{1.000  }} &  \textit{\textcolor{blue}{1.000  }} &  \textit{\textcolor{blue}{0.999  }} &  \textit{\textcolor{blue}{1.000  }} \\  
{hartmann6} & 0.957   &  0.966   &  0.964   &  0.970   &  0.965   &  0.974   &  0.970   &  \textit{\textcolor{blue}{0.976  }} &  0.974   &  \textbf{0.978  } &  0.971   \\  
{eggholder} & \textit{\textcolor{blue}{0.605  }} &  \textit{\textcolor{blue}{0.606  }} &  \textit{\textcolor{blue}{0.589  }} &  \textit{\textcolor{blue}{0.603  }} &  \textit{\textcolor{blue}{0.612  }} &  \textit{\textcolor{blue}{0.649  }} &  \textit{\textcolor{blue}{0.638  }} &  0.554   &  \textit{\textcolor{blue}{0.620  }} &  \textit{\textcolor{blue}{0.600  }} &  \textbf{0.651  } \\  
{dropwave} & 0.455   &  0.489   &  0.524   &  0.475   &  \textit{\textcolor{blue}{0.599  }} &  \textit{\textcolor{blue}{0.538  }} &  0.550   &  0.435   &  \textit{\textcolor{blue}{0.613  }} &  0.448   &  \textbf{0.651  } \\  
{beale} & \textit{\textcolor{blue}{0.920  }} &  \textit{\textcolor{blue}{0.903  }} &  \textit{\textcolor{blue}{0.910  }} &  \textbf{0.935  } &  \textit{\textcolor{blue}{0.915  }} &  \textit{\textcolor{blue}{0.927  }} &  0.874   &  \textit{\textcolor{blue}{0.901  }} &  0.902   &  \textit{\textcolor{blue}{0.912  }} &  0.900   \\  
{shubert} & 0.323   &  0.299   &  \textit{\textcolor{blue}{0.440  }} &  \textit{\textcolor{blue}{0.387  }} &  \textbf{0.551  } &  0.382   &  \textit{\textcolor{blue}{0.500  }} &  \textit{\textcolor{blue}{0.464  }} &  0.371   &  0.285   &  \textit{\textcolor{blue}{0.458  }} \\  
{sixhumpcamel6} & \textit{\textcolor{blue}{0.996  }} &  0.994   &  0.992   &  \textit{\textcolor{blue}{0.994  }} &  0.991   &  \textbf{0.997  } &  0.990   &  \textit{\textcolor{blue}{0.995  }} &  0.988   &  0.990   &  0.992   \\  
{holder} & \textit{\textcolor{blue}{0.936  }} &  0.873   &  \textit{\textcolor{blue}{0.913  }} &  \textit{\textcolor{blue}{0.941  }} &  \textit{\textcolor{blue}{0.930  }} &  \textbf{0.965  } &  \textit{\textcolor{blue}{0.949  }} &  0.950   &  \textit{\textcolor{blue}{0.948  }} &  \textit{\textcolor{blue}{0.883  }} &  \textit{\textcolor{blue}{0.936  }} \\  
{threehumpcamel} & \textbf{0.988  } &  \textit{\textcolor{blue}{0.981  }} &  \textit{\textcolor{blue}{0.978  }} &  \textit{\textcolor{blue}{0.970  }} &  0.978   &  \textit{\textcolor{blue}{0.981  }} &  0.949   &  0.975   &  0.931   &  0.971   &  0.930   \\  
{rastrigin2} & \textbf{0.917  } &  \textit{\textcolor{blue}{0.903  }} &  0.882   &  \textit{\textcolor{blue}{0.884  }} &  \textit{\textcolor{blue}{0.891  }} &  \textit{\textcolor{blue}{0.899  }} &  \textit{\textcolor{blue}{0.884  }} &  \textit{\textcolor{blue}{0.877  }} &  \textit{\textcolor{blue}{0.910  }} &  0.847   &  0.836   \\  
{rastrigin4} & \textit{\textcolor{blue}{0.806  }} &  0.759   &  0.773   &  \textit{\textcolor{blue}{0.830  }} &  \textbf{0.838  } &  \textit{\textcolor{blue}{0.834  }} &  \textit{\textcolor{blue}{0.815  }} &  0.769   &  \textit{\textcolor{blue}{0.800  }} &  0.766   &  0.775   \\  
{ackley2} & 0.850   &  0.772   &  0.838   &  0.802   &  \textbf{0.918  } &  0.832   &  \textit{\textcolor{blue}{0.869  }} &  \textit{\textcolor{blue}{0.774  }} &  0.783   &  0.811   &  \textit{\textcolor{blue}{0.896  }} \\  
{ackley5} & 0.528   &  0.557   &  0.555   &  0.579   &  0.562   &  0.602   &  0.594   &  0.604   &  \textit{\textcolor{blue}{0.620  }} &  \textbf{0.671  } &  0.621   \\  
{levy2} & 0.925   &  \textit{\textcolor{blue}{0.949  }} &  \textit{\textcolor{blue}{0.927  }} &  \textit{\textcolor{blue}{0.933  }} &  0.915   &  0.960   &  0.961   &  \textit{\textcolor{blue}{0.958  }} &  0.913   &  \textbf{0.963  } &  0.929   \\  
{levy3} & \textit{\textcolor{blue}{0.960  }} &  0.948   &  \textit{\textcolor{blue}{0.962  }} &  \textit{\textcolor{blue}{0.954  }} &  \textit{\textcolor{blue}{0.962  }} &  \textit{\textcolor{blue}{0.951  }} &  0.961   &  \textit{\textcolor{blue}{0.960  }} &  \textit{\textcolor{blue}{0.968  }} &  \textbf{0.969  } &  0.951   \\  
{levy4} & \textit{\textcolor{blue}{0.968  }} &  \textit{\textcolor{blue}{0.959  }} &  \textit{\textcolor{blue}{0.970  }} &  \textit{\textcolor{blue}{0.970  }} &  \textit{\textcolor{blue}{0.974  }} &  \textit{\textcolor{blue}{0.962  }} &  \textit{\textcolor{blue}{0.950  }} &  \textit{\textcolor{blue}{0.976  }} &  \textbf{0.976  } &  \textit{\textcolor{blue}{0.970  }} &  \textit{\textcolor{blue}{0.972  }} \\  
{griewank2} & \textit{\textcolor{blue}{0.960  }} &  \textit{\textcolor{blue}{0.963  }} &  0.952   &  \textit{\textcolor{blue}{0.958  }} &  \textbf{0.966  } &  \textit{\textcolor{blue}{0.954  }} &  0.955   &  \textit{\textcolor{blue}{0.962  }} &  \textit{\textcolor{blue}{0.958  }} &  \textit{\textcolor{blue}{0.961  }} &  \textit{\textcolor{blue}{0.960  }} \\  
{griewank5} & 0.981   &  0.984   &  0.983   &  \textit{\textcolor{blue}{0.985  }} &  \textit{\textcolor{blue}{0.984  }} &  \textit{\textcolor{blue}{0.985  }} &  0.983   &  \textbf{0.986  } &  \textit{\textcolor{blue}{0.984  }} &  \textit{\textcolor{blue}{0.985  }} &  0.983   \\  
{stybtang2} & 0.999   &  0.970   &  \textit{\textcolor{blue}{0.999  }} &  \textit{\textcolor{blue}{1.000  }} &  0.999   &  0.999   &  0.999   &  0.999   &  0.992   &  \textbf{1.000  } &  0.999   \\  
{stybtang4} & \textbf{0.937  } &  \textit{\textcolor{blue}{0.911  }} &  0.897   &  \textit{\textcolor{blue}{0.916  }} &  0.884   &  \textit{\textcolor{blue}{0.915  }} &  0.901   &  0.900   &  0.908   &  0.893   &  0.883   \\  
{powell4} & \textit{\textcolor{blue}{0.976  }} &  \textit{\textcolor{blue}{0.965  }} &  \textit{\textcolor{blue}{0.973  }} &  \textit{\textcolor{blue}{0.975  }} &  \textit{\textcolor{blue}{0.972  }} &  \textit{\textcolor{blue}{0.977  }} &  0.965   &  \textbf{0.978  } &  \textit{\textcolor{blue}{0.971  }} &  \textit{\textcolor{blue}{0.966  }} &  0.957   \\  
{dixonprice2} & \textit{\textcolor{blue}{0.988  }} &  \textit{\textcolor{blue}{0.985  }} &  \textbf{0.990  } &  \textit{\textcolor{blue}{0.989  }} &  \textit{\textcolor{blue}{0.963  }} &  \textit{\textcolor{blue}{0.967  }} &  \textit{\textcolor{blue}{0.953  }} &  \textit{\textcolor{blue}{0.959  }} &  \textit{\textcolor{blue}{0.945  }} &  \textit{\textcolor{blue}{0.982  }} &  0.953   \\  
{dixonprice4} & \textit{\textcolor{blue}{0.987  }} &  \textit{\textcolor{blue}{0.986  }} &  0.985   &  \textit{\textcolor{blue}{0.958  }} &  0.981   &  \textit{\textcolor{blue}{0.982  }} &  \textit{\textcolor{blue}{0.986  }} &  \textit{\textcolor{blue}{0.982  }} &  \textit{\textcolor{blue}{0.985  }} &  \textbf{0.987  } &  0.971   \\  
{bukin} & 0.822   &  \textit{\textcolor{blue}{0.864  }} &  \textit{\textcolor{blue}{0.865  }} &  0.844   &  \textit{\textcolor{blue}{0.860  }} &  0.851   &  \textit{\textcolor{blue}{0.861  }} &  0.852   &  \textit{\textcolor{blue}{0.850  }} &  \textbf{0.885  } &  0.826   \\  
{shekel5} & 0.273   &  \textit{\textcolor{blue}{0.383  }} &  0.400   &  \textit{\textcolor{blue}{0.414  }} &  \textit{\textcolor{blue}{0.413  }} &  \textit{\textcolor{blue}{0.402  }} &  \textit{\textcolor{blue}{0.405  }} &  \textit{\textcolor{blue}{0.425  }} &  0.366   &  \textit{\textcolor{blue}{0.401  }} &  \textbf{0.439  } \\  
{shekel7} & 0.280   &  \textit{\textcolor{blue}{0.414  }} &  0.330   &  \textit{\textcolor{blue}{0.397  }} &  0.341   &  \textit{\textcolor{blue}{0.380  }} &  0.369   &  0.378   &  \textit{\textcolor{blue}{0.406  }} &  \textbf{0.445  } &  \textit{\textcolor{blue}{0.387  }} \\  
{michal2} & 0.990   &  0.999   &  0.983   &  \textit{\textcolor{blue}{0.977  }} &  \textit{\textcolor{blue}{1.000  }} &  \textit{\textcolor{blue}{1.000  }} &  \textit{\textcolor{blue}{0.982  }} &  \textit{\textcolor{blue}{0.967  }} &  \textit{\textcolor{blue}{0.984  }} &  \textbf{1.000  } &  \textit{\textcolor{blue}{0.961  }} \\  \hline 
{Average} & 0.842   &  \textit{\textcolor{blue}{0.844  }} &  \textit{\textcolor{blue}{0.850  }} &  \textit{\textcolor{blue}{0.853  }} &  \textbf{0.861  } &  \textit{\textcolor{blue}{0.859  }} &  \textit{\textcolor{blue}{0.856  }} &  \textit{\textcolor{blue}{0.850  }} &  \textit{\textcolor{blue}{0.853  }} &  \textit{\textcolor{blue}{0.851  }} &  \textit{\textcolor{blue}{0.858  }} \\  
\bottomrule
\end{tabular}

	\label{table:all_synthetic_functions}
\end{table*}

\setlength\tabcolsep{6pt}
\begin{table*}
	\small
	\centering
	\caption{Average \gap of 100 repeats on all the five ``hard'' synthetic.}
	\begin{tabular}{llllllllllll}
    \toprule
    &{Rand} & {EI} & {2.EI.s} & {3.EI.s} & {4.EI.s} & {10.EI.s} & {12.EI.s} & {2.G} & {3.G} & {2.R.10} & {3.R.3}\\\hline
    {eggholder} & 0.498   &  0.613   &  0.633   &  0.657   &  \textit{\textcolor{blue}{0.694  }} &  \textit{\textcolor{blue}{0.704  }} &  \textbf{0.738  } &  0.583   &  0.563   &  0.569   &  0.518   \\  
    {shubert} & 0.355   &  0.408   &  \textit{\textcolor{blue}{0.441  }} &  \textbf{0.507  } &  \textit{\textcolor{blue}{0.484  }} &  \textit{\textcolor{blue}{0.455  }} &  \textit{\textcolor{blue}{0.479  }} &  0.302   &  0.254   &  0.271   &  0.297   \\  
    {bukin} & 0.600   &  \textit{\textcolor{blue}{0.849  }} &  \textit{\textcolor{blue}{0.855  }} &  \textit{\textcolor{blue}{0.859  }} &  \textbf{0.865  } &  \textit{\textcolor{blue}{0.850  }} &  0.829   &  0.829   &  0.811   &  0.772   &  0.762   \\  
    {shekel5} & 0.038   &  0.286   &  0.320   &  0.343   &  \textit{\textcolor{blue}{0.344  }} &  \textit{\textcolor{blue}{0.373  }} &  \textit{\textcolor{blue}{0.358  }} &  0.265   &  0.175   &  \textbf{0.378  } &  \textit{\textcolor{blue}{0.350  }} \\  
    {shekel7} & 0.045   &  0.268   &  0.313   &  0.325   &  0.370   &  0.358   &  \textbf{0.412  } &  0.256   &  0.174   &  0.376   &  0.361   \\  \hline
    {Average} & 0.307   &  0.485   &  0.512   &  0.538   &  \textit{\textcolor{blue}{0.551  }} &  0.548   &  \textbf{0.563  } &  0.447   &  0.395   &  0.473   &  0.458   \\  
    \bottomrule
\end{tabular}
	\label{table:synthetic_with_nonmyopic_baselines}
\end{table*}

\begin{table*}
    \centering
	\caption{Average \gap of 50 repeats on real functions for all $q$.\ei variants.}
	\label{table:real_all_qei_variants}
	\small
	\begin{tabular}{llllllllllll}
		\toprule
		&{EI} & {2.EI.b} & {2.EI.s} & {3.EI.b} & {3.EI.s} & {4.EI.b} & {4.EI.s} & {6.EI.b} & {6.EI.s} & {8.EI.b} & {8.EI.s}\\\hline
		{\svm} & 0.738   &  \textit{\textcolor{blue}{0.926  }} &  \textit{\textcolor{blue}{0.913  }} &  \textit{\textcolor{blue}{0.930  }} &  \textbf{0.940  } &  \textit{\textcolor{blue}{0.914  }} &  \textit{\textcolor{blue}{0.911  }} &  \textit{\textcolor{blue}{0.892  }} &  \textit{\textcolor{blue}{0.937  }} &  \textit{\textcolor{blue}{0.929  }} &  0.834   \\  
		{\lda} & 0.956   &  \textbf{1.000  } &  \textit{\textcolor{blue}{1.000  }} &  \textit{\textcolor{blue}{0.998  }} &  \textit{\textcolor{blue}{0.996  }} &  0.996   &  \textit{\textcolor{blue}{0.993  }} &  \textit{\textcolor{blue}{0.999  }} &  0.982   &  \textit{\textcolor{blue}{0.995  }} &  \textit{\textcolor{blue}{0.995  }} \\  
		{LogReg} & 0.963   &  \textit{\textcolor{blue}{1.000  }} &  \textit{\textcolor{blue}{0.998  }} &  \textit{\textcolor{blue}{0.999  }} &  \textit{\textcolor{blue}{1.000  }} &  \textit{\textcolor{blue}{0.999  }} &  \textit{\textcolor{blue}{0.999  }} &  \textbf{1.000  } &  \textit{\textcolor{blue}{0.999  }} &  \textit{\textcolor{blue}{1.000  }} &  \textit{\textcolor{blue}{1.000  }} \\  
		{NN Boston} & \textit{\textcolor{blue}{0.470  }} &  \textit{\textcolor{blue}{0.491  }} &  \textit{\textcolor{blue}{0.467  }} &  \textit{\textcolor{blue}{0.490  }} &  \textit{\textcolor{blue}{0.478  }} &  \textit{\textcolor{blue}{0.495  }} &  \textit{\textcolor{blue}{0.460  }} &  \textit{\textcolor{blue}{0.460  }} &  \textbf{0.502  } &  \textit{\textcolor{blue}{0.455  }} &  \textit{\textcolor{blue}{0.467  }} \\  
		{NN Cancer} & \textit{\textcolor{blue}{0.665  }} &  \textit{\textcolor{blue}{0.652  }} &  0.627   &  0.625   &  \textit{\textcolor{blue}{0.654  }} &  \textit{\textcolor{blue}{0.640  }} &  \textit{\textcolor{blue}{0.686  }} &  0.625   &  \textbf{0.700  } &  0.609   &  \textit{\textcolor{blue}{0.686  }} \\  
		{Robot3d} & 0.928   &  \textit{\textcolor{blue}{0.959  }} &  \textit{\textcolor{blue}{0.960  }} &  \textit{\textcolor{blue}{0.944  }} &  \textit{\textcolor{blue}{0.962  }} &  \textit{\textcolor{blue}{0.956  }} &  0.957   &  0.960   &  \textit{\textcolor{blue}{0.962  }} &  \textbf{0.967  } &  \textit{\textcolor{blue}{0.961  }} \\  
		{Robot4d} & \textit{\textcolor{blue}{0.730  }} &  \textit{\textcolor{blue}{0.725  }} &  \textit{\textcolor{blue}{0.726  }} &  \textit{\textcolor{blue}{0.720  }} &  0.695   &  \textbf{0.764  } &  0.695   &  \textit{\textcolor{blue}{0.760  }} &  \textit{\textcolor{blue}{0.736  }} &  \textit{\textcolor{blue}{0.732  }} &  0.697   \\ \hline
		{Average} & 0.779   &  \textit{\textcolor{blue}{0.821  }} &  \textit{\textcolor{blue}{0.813  }} &  \textit{\textcolor{blue}{0.815  }} &  \textit{\textcolor{blue}{0.818  }} &  \textit{\textcolor{blue}{0.823  }} &  \textit{\textcolor{blue}{0.815  }} &  \textit{\textcolor{blue}{0.813  }} &  \textbf{0.831  } &  \textit{\textcolor{blue}{0.812  }} &  \textit{\textcolor{blue}{0.806  }} \\  
		\bottomrule
	\end{tabular}
\end{table*}

\section{Additional Bayesian Optimization Results}
In the main paper, we presented \bo results for nine synthetic functions. 
These nine functions are selected from the 31 functions shown in Table \ref{table:all_synthetic_functions}, with \gap of \ei less than 0.9.
We only run up to 10.\ei for all functions, so 12.\ei.s and 15.\ei.s are not shown.
We argue that by identifying this set of ``hard'' functions, we are able to consistently see the advantage of nonmyopic \bo methods.
In Table \ref{table:all_synthetic_functions}, we can see all variants of our method perform better than \ei on average, but other interesting patterns are weak, possibly because they are averaged out by the ``easy'' functions.

Table \ref{table:synthetic_with_nonmyopic_baselines} includes results of rollout and \glasses on  five synthetic functions, after removing four from the nine for which the optima are located in the center of the domain. 
We remove these functions because the \acro{DIRECT} optimization procedure used in our implementations of rollout and \glasses always starts evaluating exactly at the center of the domain. 
Thus the performance of these methods on benchmarks where the global optimum just happens to be in the center is artificially inflated.
This artifact was also pointed out in \citet{lam2016bayesian}; \citet{wu2019practical} also excluded such results because of this.

We surprisingly see rollout and \glasses perform even worse than \ei on average for these five functions. 
This is an indicator that the synthetic benchmark functions are very different than the real-world functions. 
Note that \citet{malkomes2018automating} also observed significantly different results on synthetic and real functions in their unrelated \bo experiments.

Table \ref{table:real_all_qei_variants} shows the average results of 50 repeats of \ei and both ``sampling'' and ``best'' variants of $q$.\ei on the real world functions.
Different from the results on synthetic functions, we do not see ``sampling'' being consistently better than ``best'' or the other way around.

\end{document}